\documentclass[lettersize,journal]{IEEEtran}
\usepackage{graphicx}
\usepackage{multirow}
\usepackage{booktabs}
\usepackage{xspace}
\usepackage[dvipsnames]{xcolor}
\usepackage{enumitem}
\usepackage{url}
\usepackage{tikz}
\usepackage{pgfplots}
\usepackage{filecontents}
\usepackage{amsmath,amsfonts}
\usepackage{amssymb} 
\usepackage{color}
\usepackage[accsupp]{axessibility}  
\usepackage{algorithmic}
\usepackage{algorithm}
\usepackage{array}
\usepackage[caption=false,font=normalsize,labelfont=sf,textfont=sf]{subfig}
\usepackage{textcomp}
\usepackage{stfloats}
\usepackage{verbatim}
\usepackage{cite}
\usepackage[colorlinks, citecolor=Periwinkle]{hyperref}
\usepgfplotslibrary{statistics}

\newcommand*{\ReadOutElement}[4]{%
    \pgfplotstablegetelem{#2}{[index]#3}\of{#1}%
    \let#4\pgfplotsretval
}

\makeatletter
\DeclareRobustCommand\onedot{\futurelet\@let@token\@onedot}
\def\@onedot{\ifx\@let@token.\else.\null\fi\xspace}
\def\eg{\emph{e.g}\onedot} 
\def\ie{\emph{i.e}\onedot}

\def\vs{\emph{vs}\onedot}

\def\etal{\emph{et al}\onedot}
\makeatother

\newcommand{\Posecode}{Posecode\xspace}
\newcommand{\Posecodes}{Posecodes\xspace}
\newcommand{\posecode}{posecode\xspace}
\newcommand{\posecodes}{posecodes\xspace}
\newcommand{\dname}{PoseScript\xspace} 

\newcommand{\myparagraph}[1]{\vspace{0.1cm}\noindent\textbf{#1}}
\newcommand{\itemACP}[1]{\noindent $\circ$ \textit{#1}}
\newcommand{\tx}[0]{\checkmark}

\IEEEpubid{\begin{minipage}{\textwidth}\ \\[12pt]
    \centering
    \copyright~2024 IEEE.
    Personal use is permitted, but republication/redistribution requires IEEE permission.\\
    See https://www.ieee.org/publications/rights/index.html for more information. \\
    Citation information: DOI 10.1109/TPAMI.2024.3407570
\end{minipage}}

\begin{document}

\title{PoseScript: Linking 3D Human Poses and\\ Natural Language}

\author{Ginger Delmas\textsuperscript{1,2}, Philippe Weinzaepfel\textsuperscript{2}, Thomas Lucas\textsuperscript{2}, Francesc Moreno-Noguer\textsuperscript{1}, Gr\'egory Rogez\textsuperscript{2} \\[0.5cm]
\textsuperscript{1} Institut de Robòtica i Informàtica Industrial, CSIC-UPC, Barcelona, Spain\\
\textsuperscript{2} NAVER LABS Europe\\
 \textsuperscript{1}{\tt\small \{gdelmas, fmoreno\}@iri.upc.edu}, \textsuperscript{2}{\tt\small \{name.surname\}@naverlabs.com}
}



\maketitle

\begin{abstract}
Natural language plays a critical role in many computer vision applications,  such as image captioning, visual question answering, and cross-modal retrieval, to provide fine-grained semantic information. 
Unfortunately,  while human pose is key to human understanding, current 3D human pose datasets lack detailed language descriptions. To address this issue, we have introduced the PoseScript dataset. This dataset pairs more than six thousand 3D human poses from AMASS with rich human-annotated descriptions of the body parts and their spatial relationships. Additionally, to increase the size of the dataset to a scale that is compatible with data-hungry learning algorithms, we have proposed an elaborate captioning process that generates automatic synthetic descriptions in natural language from given 3D keypoints. This process extracts low-level pose information, known as ``\textit{posecodes}'', using a set of simple but generic rules on the 3D keypoints. These posecodes are then combined into higher level textual descriptions using syntactic rules. With automatic annotations, the amount of available data significantly scales up (100k), making it possible to effectively pretrain deep models for finetuning on human captions.
To showcase the potential of annotated poses, we present three multi-modal learning tasks that utilize the PoseScript dataset. Firstly, we develop a pipeline that maps 3D poses and textual descriptions into a joint embedding space, allowing for cross-modal retrieval of relevant poses from large-scale datasets. Secondly, we establish a baseline for a text-conditioned model generating 3D poses. Thirdly, we present a learned process for generating pose descriptions. These applications demonstrate the versatility and usefulness of annotated poses in various tasks and pave the way for future research in the field. The dataset is available at \url{https://europe.naverlabs.com/research/computer-vision/posescript/}.
\end{abstract}

\begin{IEEEkeywords}
3D human pose, natural language, multi-modal learning,  description, generation, retrieval, captioning. 
\end{IEEEkeywords}

\begin{figure}[t]
    \centering
    \includegraphics[width=1.0\linewidth]{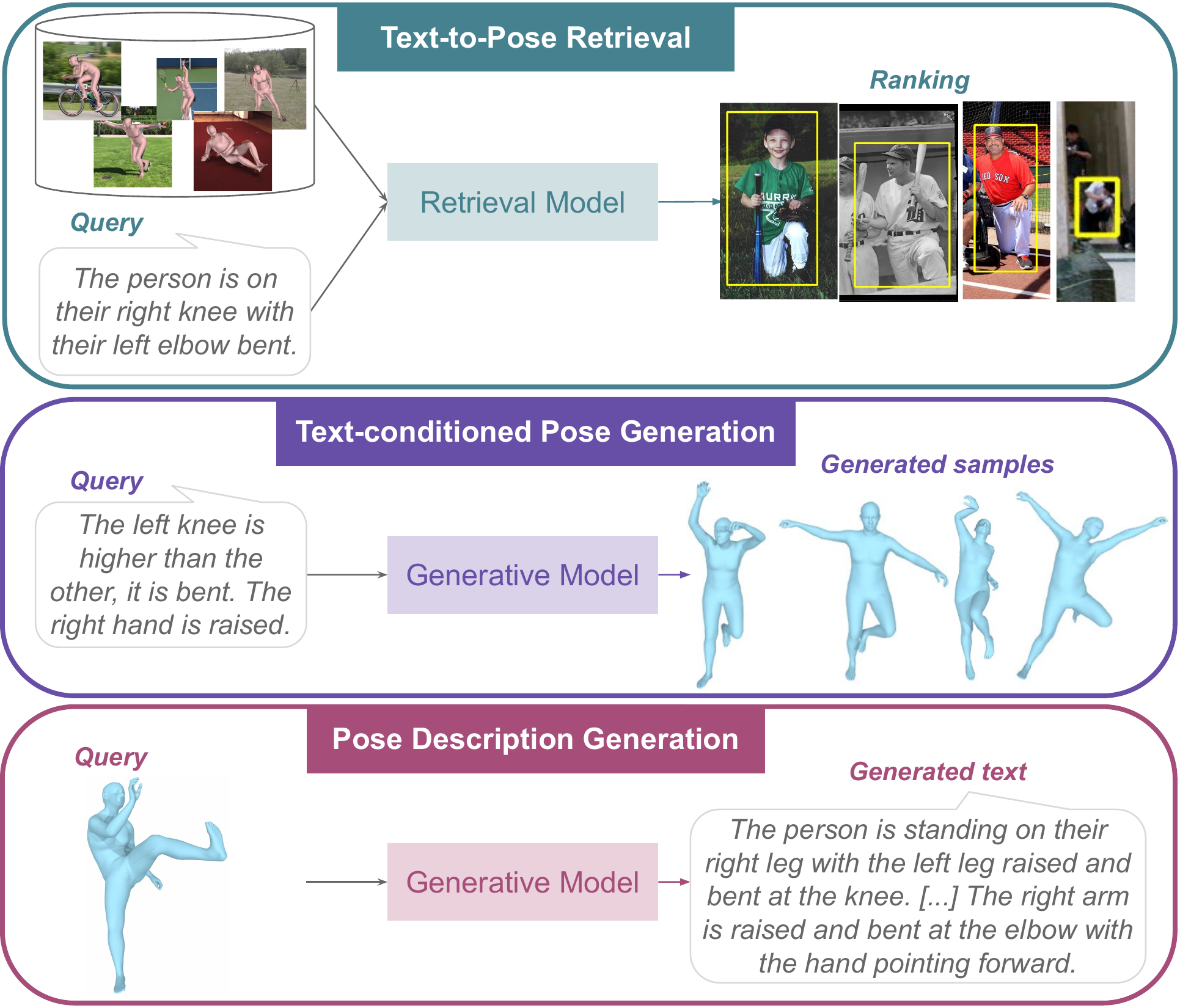} 
    \caption{\textbf{Illustration of three multi-modal learning applications that can be implemented using \dname.} The top figure illustrates text-to-pose retrieval where the goal is to retrieve poses in a large-scale database given a text query. This can be applied to databases of images with associated SMPL fits. The middle figure shows an example of text-conditioned pose generation. The bottom figure presents the generation of pose descriptions.}
    \label{fig:splash}
\end{figure}

\section{Introduction}
\label{sec:intro}

\textit{`The pose has the head down, ultimately touching the floor, with the weight of the body on the palms and the feet. The arms are stretched straight forward, shoulder width apart; the feet are a foot apart, the legs are straight, and the hips are raised as high as possible.}'.
The text above describes the downward dog yoga pose\footnote{\url{https://en.wikipedia.org/wiki/Downward_Dog_Pose}}, and a reader can easily visualize such a pose from this natural language description.
Being able to automatically map natural language descriptions and accurate 3D human poses would open the door to a number of applications such as helping image annotation when the deployment of Motion Capture (MoCap) systems is not practical; performing  {\em pose-based} semantic searches in large-scale datasets (see Figure~\ref{fig:splash} top), which are currently only based on high-level metadata such as the action being performed~\cite{h36m_pami,amass,babel}; complex pose or motion data generation in digital animation (see Figure~\ref{fig:splash} middle); or teaching posture skills to visually impaired~\cite{suveren2018teaching} (see Figure~\ref{fig:splash} bottom).
\IEEEpubidadjcol 

While the problem of combining language and images or videos has attracted significant attention~\cite{Karpathy_2015_CVPR,Vinyals2015ShowAT,Li_2017_ICCV,cross_modal_retrieval_Feng}, in particular with the impressive results obtained by the recent multimodal neural networks CLIP~\cite{clip} and DALL-E~\cite{dalle}, the problem of linking text and 3D geometry is only starting to grow. 
There have been a few recent attempts at mapping text to rigid 3D shapes~\cite{Text2Shape, poole2022dreamfusion}, and at using natural language for 3D object localization~\cite{ScanRefer} or 3D object differentiation~\cite{ShapeGlot}. More recently, Fieraru \etal~\cite{aifit} introduce AIFit, an approach to automatically generate human-interpretable feedback on the difference between a reference and a target motion.
There have also been a number of attempts to model humans using various forms of text.
Attributes have been used for instance to model body shape~\cite{bodytalk} and face images~\cite{talk2edit}.
Other approaches~\cite{Ghosh_2021_ICCV,Text2Action,petrovich21actor,Ahuja2019Language2PoseNL} leverage textual descriptions to generate motion, but without fine-grained control of the body limbs. More related to our work, Pavlakos \etal~\cite{ordinaldepth} exploit the relation between two joints along the depth dimension, and Pons-Moll \etal.~\cite{posebits} describe 3D human poses through a series of {\em posebits}, which are binary indicators for different types of questions such as `Is the right hand above the hips?'. However, these types of Boolean assertions have limited expressivity and remain far from the natural language descriptions a human would use.

\begin{figure*}[t]
    \centering
    \includegraphics[width=\textwidth]{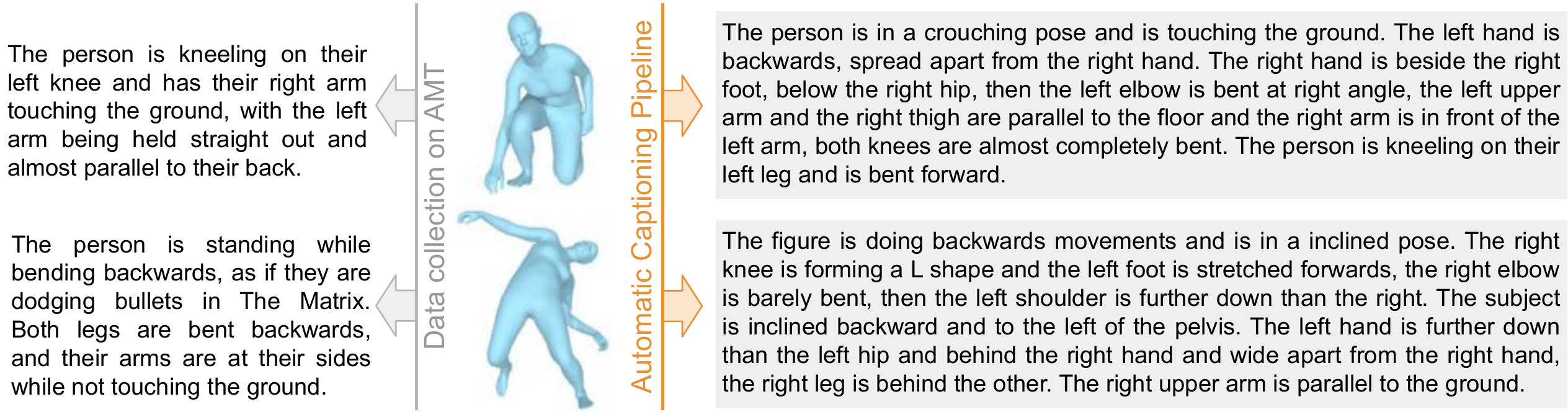}  
    \caption{\textbf{Examples of pose descriptions from \dname}, produced by human annotators (left) and by our automatic captioning pipeline (right).}
    \label{fig:captioning_examples}
\end{figure*}

In this paper, we propose to map 3D human poses with both free-form and arbitrarily complex structural descriptions, in natural language, of the body parts and their spatial relationships.
To that end, we first introduce the multi-modal {\em \dname} dataset, which consists of captions written by human annotators for about 6,000 poses from the AMASS dataset~\cite{amass}. To scale up this dataset, we additionally propose  an automatic captioning pipeline for human-centric poses that makes it possible to annotate thousands of human poses in a few minutes.
Our pipeline is built on (a) low-level information obtained via an extension of posebits~\cite{posebits} to finer-grained categorical relations of the different body parts (\eg `the knees are slightly/relatively/completely bent'), units that we refer to as {\em \posecodes},
and on (b) higher-level concepts that come either from the action labels annotated by the BABEL dataset~\cite{babel}, or combinations of \posecodes. 
We define rules to select and aggregate \posecodes using linguistic aggregation principles, and convert them into sentences to produce textual descriptions. 
As a result, we are able to automatically extract human-like captions for a normalized input 3D pose. Importantly, since the process is randomized, we can generate several descriptions per pose, as different human annotators would do.
We used this procedure to describe  100,000 poses extracted from the AMASS dataset. 
Figure~\ref{fig:captioning_examples} shows examples of human-written and automatic captions.

Using the \dname dataset, we propose to tackle three multi-modal learning tasks, see Figure~\ref{fig:splash}. 
The first is a cross-modal retrieval task where the goal is to retrieve from a database the 3D poses that are most similar to a given text query; this can also be applied to RGB images by associating them with 3D human fits.
The second task consists in generating diverse human poses conditioned on a textual description.
The third aims to generate a textual description from a provided 3D pose.
In all cases, our experiments demonstrate that it is beneficial to pretrain models using the automatic captions before finetuning them on real captions.

\noindent In summary, our contributions are  threefold:
\begin{itemize}
    \item[$\circ$] We introduce the \dname dataset (Section~\ref{sec:dataset}). It associates human poses and structural  descriptions  in  natural  language, either obtained through human-written annotations or using our automatic captioning pipeline.
    \item[$\circ$] We then study the task of text-to-pose retrieval (Section~\ref{sec:retrieval}).
    \item[$\circ$] We present the task of text-conditioned pose generation (Section~\ref{sec:generation}).
    \item[$\circ$] We finally focus on the task of pose description generation (Section~\ref{sec:text_generation}).
\end{itemize}

\noindent A preliminary version of this work was presented in~\cite{posescript2022}, in which we already introduced the PoseScript dataset and baselines for the two first tasks. In this work, we extend our contribution by increasing the size of the PoseScript dataset, studying the pose description generation problem and providing additional analysis. In particular, we study the impact of different aspects of our captioning pipeline and of the number of automatically generated captions used for pretraining. We further improved our models using the additional data, and transformers~\cite{transformer} for our text encoder.

\section{Related Work}
\label{sec:related}

\noindent \textbf{Text for humans in images.}
Some previous works have used attributes as semantic-level representation to edit body shapes~\cite{bodytalk}, human clothing~\cite{fu2022styleganhuman} and images of faces~\cite{talk2edit}. In contrast, we leverage natural language, which has the advantage of being unconstrained and more flexible. While others also do so for videos of faces~\cite{yu2022celebvtext} or images of clothed humans in modeling poses~\cite{jiang2022text2human}, our approach focuses on diverse 3D body poses and pose semantics.
Closer to our work,~\cite{zhang2020adversarial,briq2021towards} focus on generating human 2D poses, SMPL parameters or even images from captions. However, they use MS Coco~\cite{coco} captions, which are generally simple image-level statements on the activity performed by the human, and which sometimes relate to the interaction with other elements from the scene, \eg `A soccer player is running while the ball is in the air'.
By contrast, we focus on fine-grained detailed captions about the pose only.
FixMyPose~\cite{fixmypose} provides manually annotated captions about the difference between human poses in two synthetic images. These captions also include information abouth the objects in the environment, \eg `carpet' or `door'. Similarly, AIFit~\cite{aifit} proposes to automatically generate text about the discrepancies between a reference motion and a performed one, based on differences of angles and positions.
Our approach, instead, is focused on describing one single pose without relying on any other visual element.

\myparagraph{Text for human motion.}
We deal with  static poses, whereas several existing methods have 
mainly studied 3D action (sequence) recognition or text-based 2D~\cite{Text2Action} or 3D motion synthesis. These approaches either condition their model on action labels \cite{chuan2020action2motion,petrovich21actor,posegpt}, or descriptions in natural language~\cite{Plappert2018LearningAB,Lin2018GeneratingAV, PairedRecurrentAutoencoders, Ahuja2019Language2PoseNL, Ghosh_2021_ICCV, petrovich2022temos, Guo_2022_CVPR, tevet2022motionclip, chuan2022tm2t, kim2022flame, tevet2023human, zhang2022motiondiffuse}. Alternatively, other works leverage appearance information~\cite{hong2022avatarclip, youwang2022clipactor}.
Yet, even if motion descriptions effectively constrain \textit{sequences} of poses, they do not specifically inform about individual poses. What if an animation studio looks for a sequence of 3D body poses where `the man is running with his hands on his hips'? The model used by the artists to initialize the animation should have a deep understanding of the relations between the body parts. To this end, it is important to learn about specific pose semantics, beyond global pose sequence semantics.

\myparagraph{Pose semantic representations.}
Our captioning generation process relies on \posecodes that capture relevant information about the pose semantics.
\Posecodes are draw inspiration on posebits~\cite{posebits}, where images showing a human are annotated with various binary indicators. This data is then used to reduce ambiguities in 3D pose estimation. Conversely, we automatically extract \posecodes from normalized 3D poses in order to generate descriptions in natural language. Ordinal depth~\cite{ordinaldepth} can be seen as a special case of posebits, focusing on the depth relationship between two joints. The authors exploit such annotations on training images to improve a human mesh recovery model by adding extra constraints. Poselets~\cite{poselets} can also be seen as another way to extract discriminative pose information, but they are not easily interpreted. In contrast to these representations, we propose to generate pose descriptions in natural language, which have the advantage (a) of being a very intuitive way to communicate ideas, and (b) of providing greater flexibility.
More recently, TIPS~\cite{Roy_2022_ECCV_tips} introduce structural descriptions for poses from DeepFashion~\cite{liu2016deepfashion} images. These are collected by selecting states for some body parts among predefined lists. While the underlying process of our proposed captioning pipeline is similar, we  deal with 3D poses instead of images, and devise an automatic approach. Besides, we do not limit to fashion and focus on a large variety of poses. Eventually, we collect actual free-form text descriptions from human annotators.

In summary, our proposed \dname dataset differs from existing datasets in that it focuses on single 3D poses instead of motion~\cite{plappert2016kit}, which are diverse and not restricted to modeling poses~\cite{Roy_2022_ECCV_tips}. Moreover, it provides direct descriptions in natural language instead of simple action labels~\cite{babel,chuan2020action2motion,shahroudy2016ntu,li2010action,h36m_pami}, binary relations~\cite{posebits,ordinaldepth} or modifying texts~\cite{fixmypose,aifit}.
\begin{figure}[t!]
 \centering
  \includegraphics[width=0.8\linewidth]{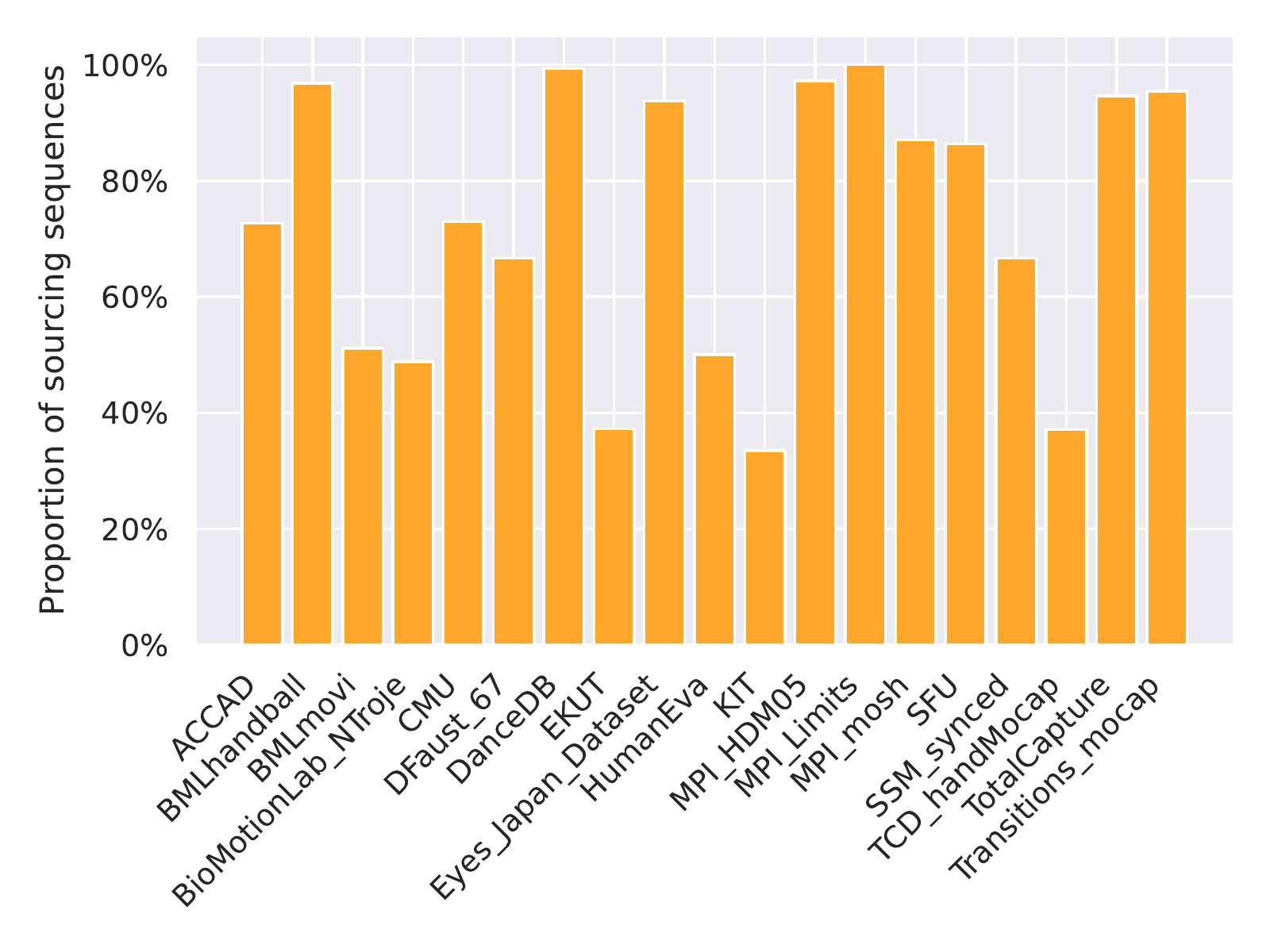}
 \includegraphics[width=0.8\linewidth]{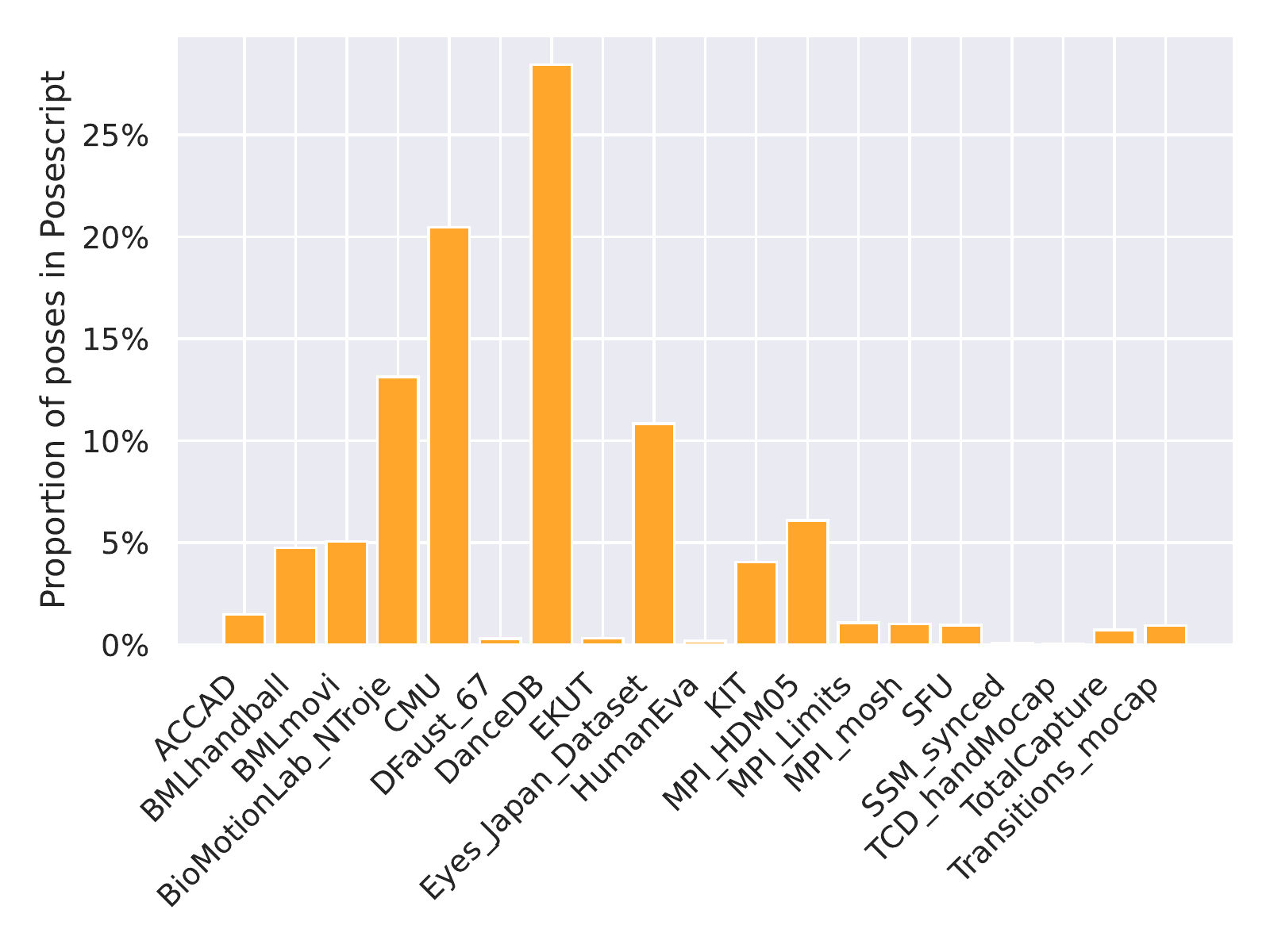}
 \vspace{-5mm}
 \caption{\textbf{Origin of the selected poses.} The top bar plot shows the proportion of sequences that are eventually used in \dname with respect to available sequences in AMASS. A sequence is `used' if it provided at least one pose to \dname. The bottom bar plot shows the distribution of the \dname poses over the AMASS sub-datasets.}
 \label{fig:subdataset_origin}
\end{figure}

\section{The \dname Dataset}
\label{sec:dataset}

The \dname dataset is composed of static 3D human poses, together with fine-grained semantic annotations in natural language. 
We provide \textbf{H}uman-written annotated descriptions (\dname-H), and further increase the amount of data with \textbf{A}utomatically generated captions (\dname-A).
The pose selection strategy is presented in Section~\ref{sec:pose_selection}, the crowd-sourced data collection process in Section~\ref{sub:process}, and the automatic captioning pipeline in Section~\ref{sub:automatic}.
Finally, aggregated statistics over the \dname dataset are reported in Section~\ref{sub:stats}.

\subsection{Pose selection}
\label{sec:pose_selection}

The \dname dataset contains a total of 100,000 human poses sampled from 14,096 AMASS~\cite{amass} sequences.  
To ensure diversity, we excluded the first and last 25 frames of each sequence, which contained initialization poses such as T-poses. Additionally, we sampled only one pose from every 25 frames to avoid redundant poses that were too similar to each other. To further maximize variability, we utilized a farthest-point sampling algorithm. First, we normalized the joint positions of each pose using the neutral body model with default shape coefficients and global orientation set to 0. Then, we randomly selected one pose from the dataset and iteratively added the pose with the highest MPJE (mean per-joint error) to the set of already-selected poses. This process continued until we obtained the desired number of poses with maximum variability.

Figure~\ref{fig:subdataset_origin} presents the AMASS sub-datasets from which the 100,000 selected poses were chosen. Notably, it appears that all sequences of DanceDB and MPI\_Limits available in AMASS are used in \dname; and that most of the poses in \dname actually come from DanceDB (28\%), CMU (20\%) and BioMotionLab (13\%). This is because these sub-datasets exhibit a greater variability of poses compared to  others in AMASS.

\subsection{Dataset collection}
\label{sub:process}

We collect human-written captions for 3D human poses extracted from the AMASS dataset~\cite{amass}, using Amazon Mechanical Turk\footnote{\url{https://www.mturk.com}} (AMT), a crowd-sourced annotation platform.

\begin{figure}[t!]
    \centering
    \includegraphics[width=\linewidth]{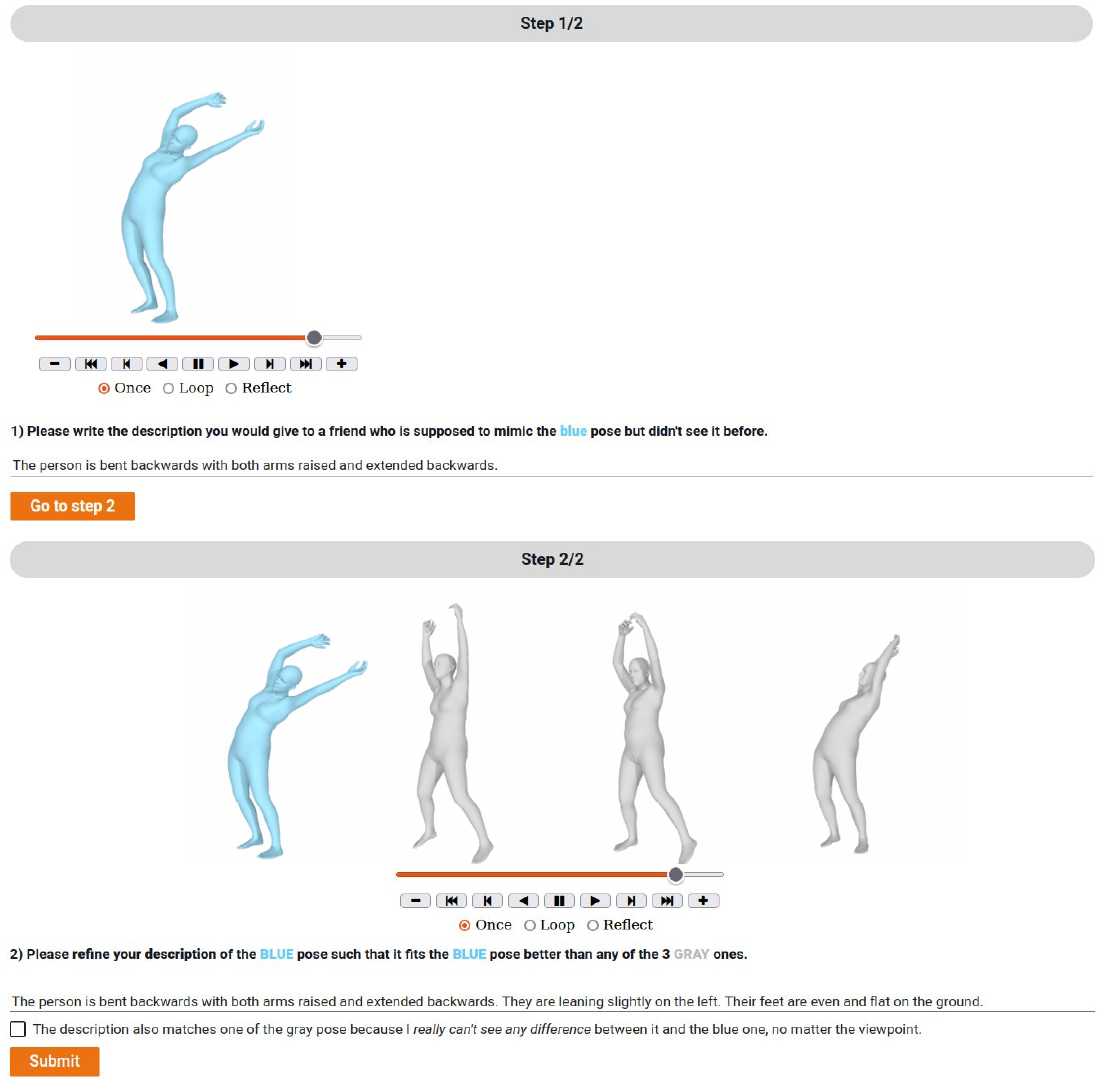} \\[-0.3cm]
    \caption{\textbf{Interface presented to the AMT annotators} in order to collect discriminative descriptions of the blue pose following a two-step process.}
    \label{fig:amt_interface}
\end{figure}

\myparagraph{Annotation interface.}
The interface, depicted in Figure~\ref{fig:amt_interface}, first presents the annotators with the mesh of a human pose (in blue), and a slider for controlling the viewpoint. The task is to write the description of the blue pose, using directions relative to the subject (the `\textit{left}' is the body's `\textit{left}'), inter- body parts indications (\eg `\textit{the right hand is on the hip}'), common pose references (\eg `\textit{in a headstand}') and analogies. In a second step, to encourage discriminative captions, we additionally display 3 discriminator poses (in gray), which are semantically close to the pose to annotate. The workers are then instructed to refine their description such that it fits only the blue pose. While this interface was first designed to be one-step (with all poses shown at once), we found that annotators would sometimes just provide enough information about the blue pose to distinguish it from the displayed gray poses, but not enough information to fully describe it in detail. This two-step design is an attempt at limiting this phenomenon, in order to obtain both complete and precise descriptions. Some \dname-H examples are shown in Figure~\ref{fig:captioning_examples} (left).

\myparagraph{Pose discriminators} are selected in PoseScript.
They should be similar to the target pose, as measured by the distance between pose embeddings obtained with an early version of our retrieval model. They are also required to have at least 15 different posecode categorizations. This ensures that the selected pose discriminators share some semantic similarities with the pose to be annotated while having sufficient differences to be easily  distinguished by the annotators.

\myparagraph{Annotators qualifications.}
The annotation task was initially made available to workers living in English-speaking countries, who got at least 5000 of their previous assignments approved, and have an approval rate over 95\%. We manually reviewed close to 1000 annotations, based on the following criteria: the description is `complete' (\ie, nearly all the body parts are described), there is no left/right confusion, no distance metric (as these are not scalable with the body size), no subjective comments, few spelling errors and good grammar.
Based on these evaluations, we further qualified 41 workers and made the assignments available to them only; only some of their annotations, selected at random, would then be manually reviewed. 61 other workers submitted great annotations, but did not complete enough to get qualified.

\myparagraph{Pricing.}
The time to complete a HIT was estimated to be 2-3 minutes. Each HIT was rewarded \$0.50, based on the minimum wage in California for 2022. We additionally paid \$2 bonus to qualified annotators for every 50 annotations.

\myparagraph{Semantic analysis.}
We report in Table~\ref{tab:semantic_analysis} the results of a semantic analysis carried out on 115 annotations. It emerges that one challenging aspect of the dataset is the implicit side description of some body parts: the deduction of the corresponding side involves reasoning about previously described body parts and hierarchical relations between them. In the course of this study, we moreover measured that the annotations describe in average 6.2 different body parts: those vary across annotations, and multiple indications can be given to detail the position of one body part. This hints at the level of details in the annotations.

\begin{table*}[t]
    \caption{\textbf{Semantic analysis} on 115 PoseScript annotations.}
    \centering
    \resizebox{0.8\textwidth}{!}{%
    \begin{tabular}{lcc}
    \toprule
    Property & Proportion & Examples \\
    \midrule
    Egocentric relations & 86\% & ``\textit{above their head}'' / ``\textit{looking at their right arm}'' / ``\textit{at about shin level}'' \\
    Analogies & 24\% & ``\textit{in a circular position}'' / ``\textit{almost in a seated position}'' / ``\textit{like a hug}'' \\
    Implicit side description & 41\% & ``\textit{right leg stretched out forward, $\O$ heel on the ground}'' \\
    \bottomrule
    \end{tabular}
    }
    \label{tab:semantic_analysis}
\end{table*}

\subsection{Automatic captioning pipeline}
\label{sub:automatic}

\begin{figure*}[t!]
    \centering
    \includegraphics[width=0.95\textwidth]{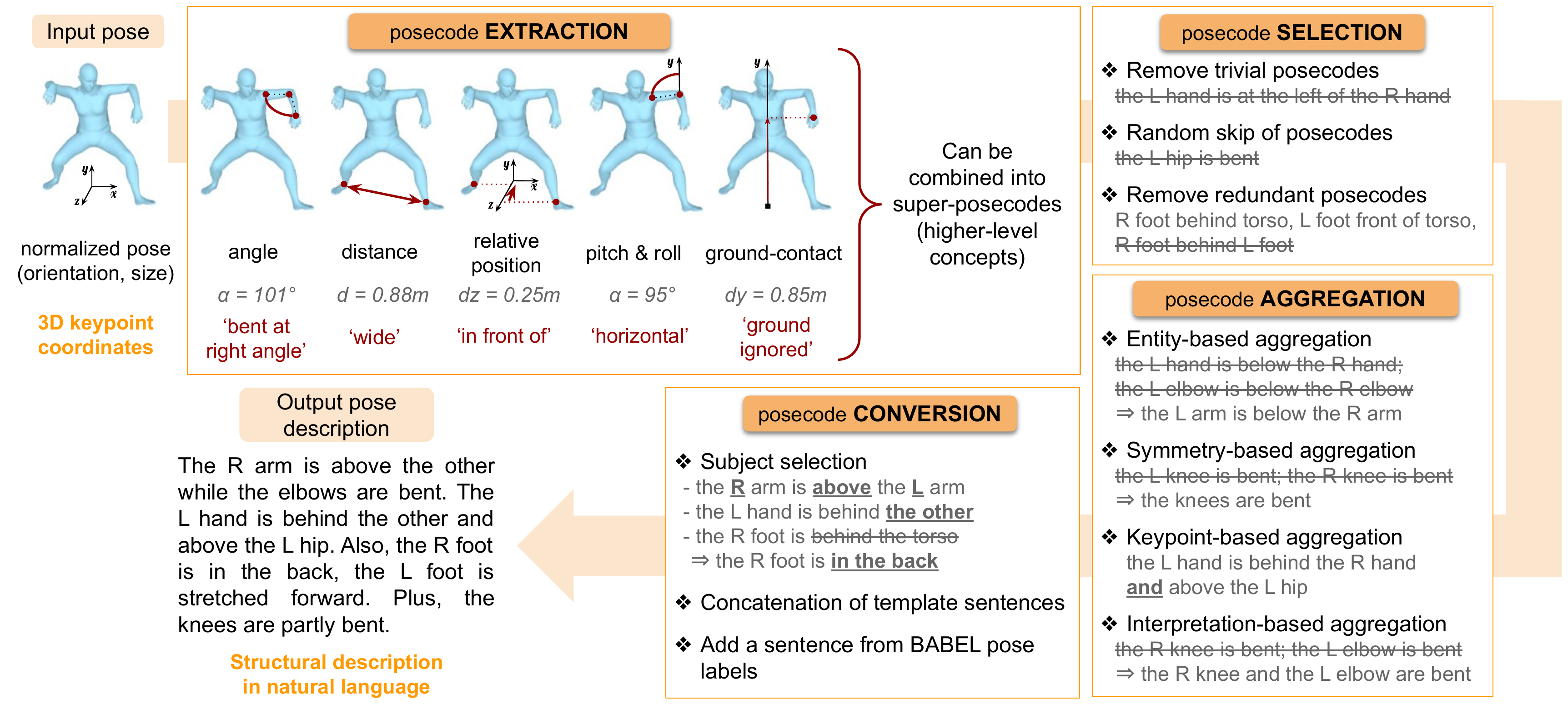} \\[-0.3cm]
    \caption{\textbf{Overview of our captioning pipeline.} Given a normalized 3D pose, we use \posecodes to extract semantic pose information. These \posecodes are then selected, merged or combined (when relevant) before being converted into a structural pose description in natural language. Letters `L' and `R' stand for `left' and `right' respectively.}
    \label{fig:captioning_pipeline}
\end{figure*}

We now describe the process used to generate synthetic textual descriptions for 3D human poses. 
As depicted in Figure~\ref{fig:captioning_pipeline}, it relies on the extraction, selection and aggregation of elementary pieces of pose information, called \textit{\posecodes}, that are eventually converted into sentences to produce a description.

The process takes 3D keypoint coordinates of human-centric poses as input. These are inferred with the SMPL-H body model~\cite{mano} using the default shape coefficients and a normalized global orientation along the y-axis.

\myparagraph{1. \Posecode extraction.}
A \posecode describes a relation between a specific set of joints.
We capture five kinds of elementary relations: angles, distances and relative positions (as in~\cite{posebits}), but also pitch, roll and ground-contacts.

\itemACP{Angle \posecodes} describe how a body part `bends' at a given joint, \eg the left elbow.
Depending on the angle, the \posecode is assigned one of the following attributes: `\texttt{straight}', `\texttt{slightly bent}', `\texttt{partially bent}', `\texttt{bent at right angle}', `\texttt{almost completely bent}' and `\texttt{completely bent}'.

\itemACP{Distance \posecodes} categorize the $L2$-distance between two keypoints (\eg the two hands) into `\texttt{close}', `\texttt{shoulder width apart}', `\texttt{spread}' or `\texttt{wide}' apart.

\itemACP{\Posecodes on relative position} compute the difference between two keypoints along a given axis.
The possible categories are, for the $x$-axis: `\texttt{at the right of}', `\texttt{x-ignored}', `\texttt{at the left of}'; for the $y$-axis: `\texttt{below}', `\texttt{y-ignored}', `\texttt{above}'; and for the $z$-axis: `\texttt{behind}', `\texttt{z-ignored}' and `\texttt{in front of}'.
In particular, comparing the $x$-coordinate of the left and right hands allows to infer if they are crossed (\ie, the left hand is `\texttt{at the right}' of the right hand). The `\texttt{ignored}' interpretations are ambiguous configurations which will not be described.

\itemACP{Pitch \& roll \posecodes} assess the verticality or horizontality of a body part defined by two keypoints (\eg the left knee and hip together define the left thigh).
A body part is `\texttt{vertical}' if it is approximately orthogonal to the $y$-hyperplane, and `\texttt{horizontal}' if it is in it.
Other configurations are `\texttt{pitch-roll-ignored}'.

\itemACP{Ground-contact \posecodes}, used for intermediate computation only, denote whether a keypoint is `\texttt{on the ground}' (\ie, vertically close to the keypoint of minimal height in the body, considered as the ground) or `\texttt{ground-ignored}'.

\noindent \textit{Handling ambiguity in \posecode categorization.} \Posecode categorizations are obtained using predefined thresholds. As these values are inherently subjective, we randomize the binning step by also defining a noise level applied to the measured angles and distances values before thresholding.

\noindent \textit{Higher-level concepts.} We also define a few \emph{super-\posecodes} to extract higher-level pose concepts. These \posecodes are binary (they either apply or not to a given pose configuration), and are expressed from elementary \posecodes. For instance, the super-\posecode `\texttt{kneeling}' can be defined as having both knees `\texttt{on the ground}' and `\texttt{completely bent}'.

\myparagraph{2. \Posecode selection} aims at selecting an interesting subset of \posecodes among those extracted, to obtain a concise yet discriminative description.
First, we remove trivial settings (\eg `the left hand is at the left of the right hand').
Next, based on a statistical study over the whole set of poses, we randomly skip a few non-essential --\ie, non-trivial but non highly discriminative -- \posecodes, to account for natural human oversights. We also set highly-discriminative \posecodes as unskippable. Finally, we remove redundant \posecodes based on statistically frequent pairs and triplets of \posecodes, and transitive relations between body parts. Details are provided in the supplementary material.

\myparagraph{3. \Posecode aggregation} consists in merging together \posecodes that share semantic information. This reduces the size of the caption and makes it more natural. We propose four specific aggregation rules:

\itemACP{Entity-based aggregation} merges \posecodes that have similar categorizations 
while describing keypoints belonging to larger entities (\eg the arm or the leg).
For instance `the left hand is below the right hand' + `the left elbow is below the right hand' is combined into `the left arm is below the right hand'.

\itemACP{Symmetry-based aggregation} fuses \posecodes that share the same categorization,
and operate on joint sets that differ only by their side of the body. The joint of interest is hence put in plural form, \eg `the left elbow is bent' + `the right elbow is bent' becomes `the elbows are bent'.

\itemACP{Keypoint-based aggregation} brings together \posecodes with a common keypoint. 
We factor the shared keypoint as the subject and concatenate the descriptions. The subject can be referred to again using \eg `it' or `they'.
For instance, `the left elbow is above the right elbow' + `the left elbow is close to the right shoulder' + `the left elbow is bent' is aggregated into `The left elbow is above the right elbow, and close to the right shoulder. It is bent.'.

\itemACP{Interpretation-based aggregation} merges \posecodes that have the same categorization,
but apply on different joint sets (that may overlap). Conversely to entity-based aggregation, it does not require that the involved keypoints belong to a shared entity. For instance, `the left knee is bent' + `right elbow is bent'  becomes `the left knee and the right elbow are bent'.

Aggregation rules are applied at random when their conditions are met. In particular, joint-based and interpretation-based aggregation rules may operate on the same \posecodes. To avoid favouring one rule over the other, merging options are first listed together and then applied at random.

\myparagraph{4. \Posecode conversion into sentences} is performed in two steps.
First, we select the subject of each \posecode. For symmetrical posecodes -- which involve two joints that only differ by their body side --  the subject is chosen at random between the two keypoints, and the other is randomly referred to by its name, its side or `the other' to avoid repetitions and provide more varied captions.
For asymmetrical \posecodes, we define a `main' keypoint (chosen as subject) and `support' keypoint, used to specify pose information (\eg the `head' in `the left hand is raised above the head').   
For the sake of flow, in some predefined cases, we omit to name the support keypoint (\eg `the left hand is raised above the head' is reduced to `the left hand is raised').
Second, we combine all \posecodes together in a final aggregation step.
We obtain individual descriptions by plugging each \posecode data into one template sentence, picked at random in the set of possible templates for a given \posecode category.
Finally, we concatenate the pieces in random order, using random pre-defined transitions. 
Optionally, for poses extracted from annotated sequences in BABEL~\cite{babel}, we add a sentence based on the associated high-level concepts (\eg `the person is in a yoga pose').

Some automatic captioning examples are presented in Figure~\ref{fig:captioning_examples} (right).
The captioning process is highly modular; it allows to simply define, select and aggregate the \posecodes based on different rules.
Design of new kinds of \posecodes (especially super-\posecodes) or additional aggregation rules, can yield further improvements in the future.
Importantly, randomization has been included at each step of the pipeline which makes it possible to generate different captions for the same pose, as a form of data augmentation.

\subsection{Dataset statistics}
\label{sub:stats}

\begin{figure}[t!]
    \centering
    \includegraphics[width=0.9\linewidth]{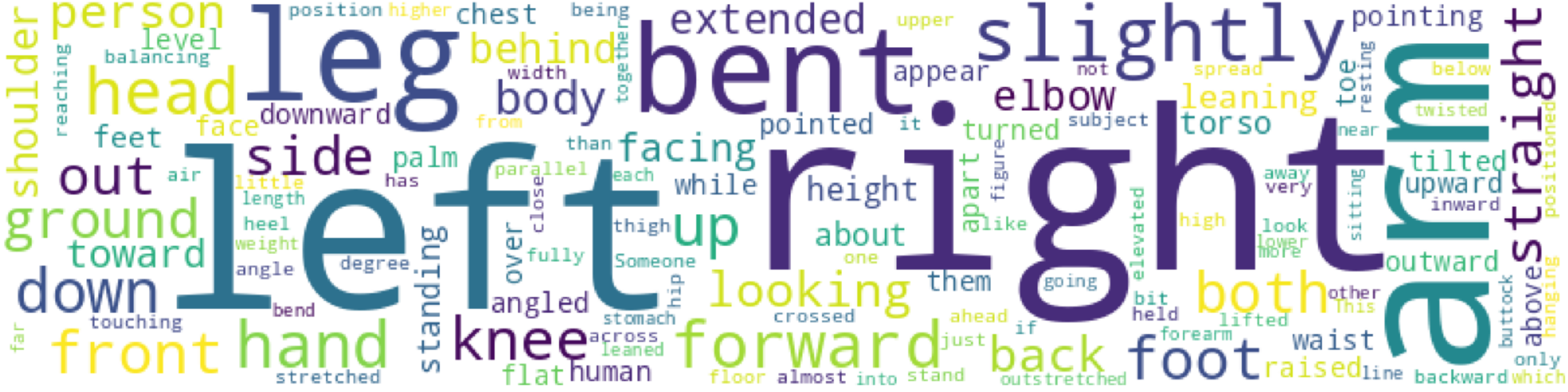} \\[-0.3cm]
    \caption{\textbf{Wordcloud} of the most frequent words in PoseScript-H descriptions.}
    \label{fig:wordcloud}
\end{figure}

We collected 6,283 human annotations on AMT (\dname-H).
We semi-automatically clean the descriptions by manually correcting the spelling of words that are not in the English dictionary, by removing one of two identical consecutive words, and by checking the error detected by a spell checker, namely NeuSpell~\cite{neuspell}. Human-written descriptions have an average length of 54.2 tokens (50.3 words, plus punctuation). An overview of the most frequent words, among a vocabulary of 1866, is presented in Figure~\ref{fig:wordcloud}. 

We used the automatic captioning pipeline to increase the number of pose descriptions in the dataset (\dname-A). We designed a total of 87 \posecodes, and automatically generated 3 captions for each of the 100,000 poses, in less than 10 minutes.  In other words, we can generate close to 1 million captions in the times it takes to manually write 10. Overall, automatic descriptions were produced using a \posecode skipping rate of 15\%, and an aggregation probability of 95\%.
Further \posecodes statistics are provided in the supplementary.

We split the dataset into roughly 70\% for training, 10\% for validation and 20\% for testing while ensuring that poses from a same AMASS sequence belong to the same split.
\section{Application to Text-to-Pose Retrieval}
\label{sec:retrieval}

This section addresses the problem of \emph{text-to-pose retrieval}, which involves ranking a large collection of poses based on their relevance to a given textual query. This task is also relevant for the inverse problem of \emph{pose-to-text retrieval}. To tackle  cross-modal retrieval problems, it is common practice to encode both modalities into a shared latent space.

\myparagraph{Problem formulation.}
Let $S = \{(c_i, p_i)\}_{i=1}^N$ be a set of caption-and-pose pairs. By construction, $p_i$ is the most relevant pose for caption $c_i$, which means that $p_{j \neq i}$ should be ranked after $p_i$ for text-to-pose retrieval. In other words, the retrieval model aims to learn a similarity function $s(c,p) \in \mathbb{R}$ such that $s(c_i, p_i) > s(c_i, p_{j \neq i})$. 
 By computing and ranking the similarity scores between the query and each pose in the collection, a set of relevant poses can be retrieved for a given text query (and vice versa for pose-to-text retrieval).

\begin{figure}[t]
    \centering
    \includegraphics[width=\linewidth]{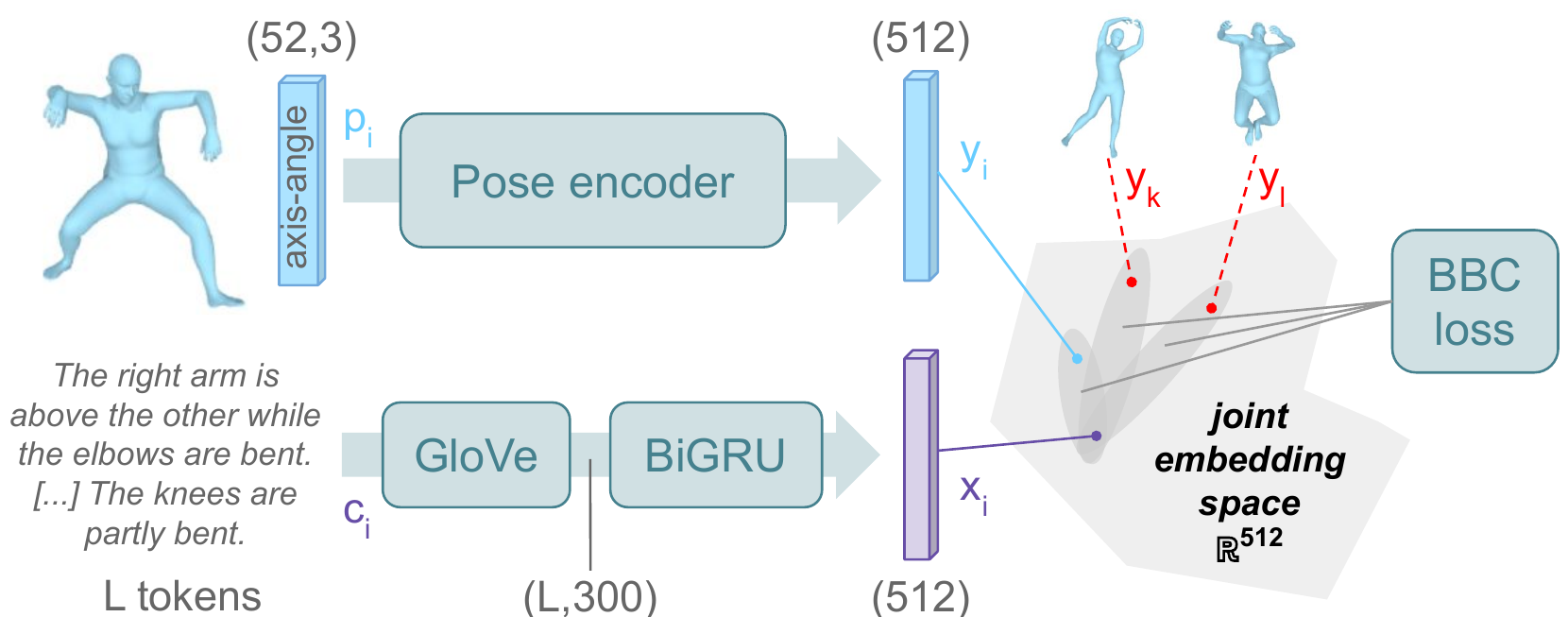} \\[-0.1cm]
    \caption{\textbf{Overview of the training scheme of the retrieval model.} The input pose and caption are first separately fed into a pose encoder and a text encoder, respectively, to map them into a joint embedding space. The loss function is designed to encourage the pose embedding $y_i$ and its corresponding caption embedding $x_i$ to be close to each other in this shared latent space, while also pulling them apart from the features of other poses in the same training batch (\eg, $y_k$ and $y_l$).
    }
    \label{fig:model}
\end{figure}

Given that poses and captions belong to different modalities, we employ separate encoders to embed them into a common latent space. Specifically, we use a textual encoder $\theta(\cdot)$ and a pose encoder $\phi(\cdot)$ to encode the captions and poses, respectively. Let $x = \theta(c) \in \mathbb{R}^d$ and $y = \phi(p) \in \mathbb{R}^d$ be the $L2$-normalized representations of a caption $c$ and a pose $p$ in the joint embedding space, as shown in Figure~\ref{fig:model}. The similarity score between the two modalities is computed based on the distance between their respective embeddings.

\myparagraph{Encoders.} The caption is tokenized and then embedded using either a bi-GRU~\cite{bigru} on top of GloVe word embeddings~\cite{pennington2014glove} or a transformer~\cite{transformer} on top of the frozen pretrained DistilBERT~\cite{sanh2019distilbert} word embeddings.
The pose is encoded as a matrix of size $(52,3)$, consisting in the rotation of the SMPL-H body joints in axis-angle representation. It is flattened and fed to the pose encoder, chosen as the VPoser encoder~\cite{smplx}. An added ReLU and final projection layer produce an embedding of the same size $d$ as the text encoding.

\myparagraph{Training.}
Given a batch of $B$ training pairs $(x_i,y_i)$, we use the Batch-Based Classification (BBC) loss which is common in cross-modal retrieval~\cite{vo2019composing}:
\begin{equation}
    \mathcal{L}_{\text{BBC}} = - \frac{1}{B} \sum_{i=1}^B \text{log} \frac{\text{exp} \big( \gamma \sigma(x_i, y_i) \big) }{\sum_j \text{exp} \big( \gamma \sigma(x_i, y_j) \big) },
\end{equation}
where $\gamma$ is a learnable temperature parameter and 
$\sigma$
is the cosine similarity function $ \sigma(x,y) = x^\top y / \big( \Vert x \Vert_2 \times \Vert y \Vert_2 \big)$.

\myparagraph{Evaluation protocol.}
Text-to-pose retrieval is evaluated by ranking the whole set of poses for each of the query texts. We then compute the recall@K ($R@K$), which is the proportion of query texts for which the corresponding pose is ranked in the top-$K$ retrieved poses. We proceed similarly to evaluate pose-to-text retrieval.
We use K = 1, 5, 10 and additionally report the mean recall (mRecall) as the average over all recall@K values from both retrieval directions.

\begin{table*}[t]
    \caption{\textbf{Text-to-pose and pose-to-text retrieval results} on the test split of the \dname dataset. For human-written captions (\dname-H), we evaluate models trained on each specific caption set alone, and one pretrained on automatic captions (\dname-A) then finetuned (FT) on human captions. Unless specified otherwise, models all have the GloVe-biGRU configuration. Results are averaged over 3 runs.}
    \centering
    \resizebox{0.9\textwidth}{!}{%
    \begin{tabular}{lccccccc@{}}
    \toprule
     &  \multirow{2}{*}{mRecall\color{OliveGreen}{$\uparrow$} } & \multicolumn{3}{c}{pose-to-text} & \multicolumn{3}{c}{text-to-pose} \\
     
    \cmidrule(lr){3-5} \cmidrule(lr){6-8}
    
     & & $R@1$\color{OliveGreen}{$\uparrow$} & $R@5$\color{OliveGreen}{$\uparrow$} & $R@10$\color{OliveGreen}{$\uparrow$} & $R@1$\color{OliveGreen}{$\uparrow$} & $R@5$\color{OliveGreen}{$\uparrow$} & $R@10$\color{OliveGreen}{$\uparrow$} \\
    \midrule
    
    \multicolumn{8}{l}{\textit{test on \dname-A (19,990 samples)}} \\
    ~~~~trained on \dname-A & \textbf{72.8} \tiny{${\pm}$ 0.4} & \textbf{47.2} \tiny{${\pm}$ 0.5} & \textbf{78.5} \tiny{${\pm}$ 0.3} & \textbf{87.1} \tiny{${\pm}$ 0.2} & \textbf{52.4} \tiny{${\pm}$ 0.8} & \textbf{82.0} \tiny{${\pm}$ 0.5} & \textbf{89.3} \tiny{${\pm}$ 0.4} \\
    \midrule
    \multicolumn{8}{l}{\textit{test on \dname-H (1234 samples)}} \\
    ~~~~trained on \dname-A & 5.9 \tiny{${\pm}$ 0.4} & 2.3 \tiny{${\pm}$ 0.4} & 6.9 \tiny{${\pm}$ 0.7} & 11.3 \tiny{${\pm}$ 0.6} & 1.4 \tiny{${\pm}$ 0.1} & 5.1 \tiny{${\pm}$ 0.3} & 8.6 \tiny{${\pm}$ 0.4} \\
    ~~~~trained on \dname-H & 23.0 \tiny{${\pm}$ 0.6} & 8.9 \tiny{${\pm}$ 0.3} & 24.4 \tiny{${\pm}$ 1.5} & 34.8 \tiny{${\pm}$ 0.5} & 9.3 \tiny{${\pm}$ 1.0} & 24.6 \tiny{${\pm}$ 0.1} & 35.7 \tiny{${\pm}$ 0.9} \\
    ~~~~trained on \dname-A, FT on \dname-H & 40.9 \tiny{${\pm}$ 0.1} & 19.8 \tiny{${\pm}$ 0.4} & 44.9 \tiny{${\pm}$ 0.7} & 56.2 \tiny{${\pm}$ 0.7} & 19.9 \tiny{${\pm}$ 0.6} & 46.5 \tiny{${\pm}$ 0.1} & 57.9 \tiny{${\pm}$ 0.3} \\
    ~~~~ ~~~~\textit{using the transformer text encoder} & 43.3 \tiny{${\pm}$ 0.6} & 21.0 \tiny{${\pm}$ 0.5} & 48.2 \tiny{${\pm}$ 1.2} & 60.4 \tiny{${\pm}$ 1.0} & 21.7 \tiny{${\pm}$ 1.3} & 48.1 \tiny{${\pm}$ 0.5} & 60.6 \tiny{${\pm}$ 0.5} \\
    ~~~~ ~~~~ ~~~~\textit{with mirroring augmentation} & \textbf{45.3} \tiny{${\pm}$ 0.4} & \textbf{22.3} \tiny{${\pm}$ 1.6} & \textbf{50.1} \tiny{${\pm}$ 1.0} & \textbf{62.9} \tiny{${\pm}$ 0.7} & \textbf{22.1} \tiny{${\pm}$ 0.4} & \textbf{51.4} \tiny{${\pm}$ 0.3} & \textbf{63.1} \tiny{${\pm}$ 0.9} \\
    \bottomrule
    \end{tabular}
    }
    \label{tab:results}
    \end{table*}

\myparagraph{Quantitative results.}
We report results on the test set of \dname in Table~\ref{tab:results}, both on automatic and human-written captions. Our model trained on automatic captions obtains a mean recall of 72.8\%, with a R@1 close to 50\% and a R@10 above 80\% on automatic captions.
However, the performance degrades on human captions, as many words from the richer human vocabulary are unseen during training on automatic captions. When trained on human captions, the model obtains a higher -- but still rather low -- performance. Using human captions to finetune the initial model trained on automatic ones brings an improvement of almost a factor 2, with a mean recall (resp. R@10 for text-to-pose) of 40.9\% (resp. 57.9\%) compared to 23.0\% (resp. 35.7\%) when training from scratch. This experiment clearly shows the benefit of using the automatic captioning pipeline to scale-up the \dname dataset. In particular, this suggests that the model is able to derive new concepts in human-written captions from non-trivial combination of existing \posecodes in automatic captions.
The last two rows show further improvement when using a transformer-based text encoder and applying data augmentation by mirroring the poses (\ie, switching \textit{left} and \textit{right} side words in the texts).

\begin{figure}[t]
    \centering
    \includegraphics[width=\linewidth]{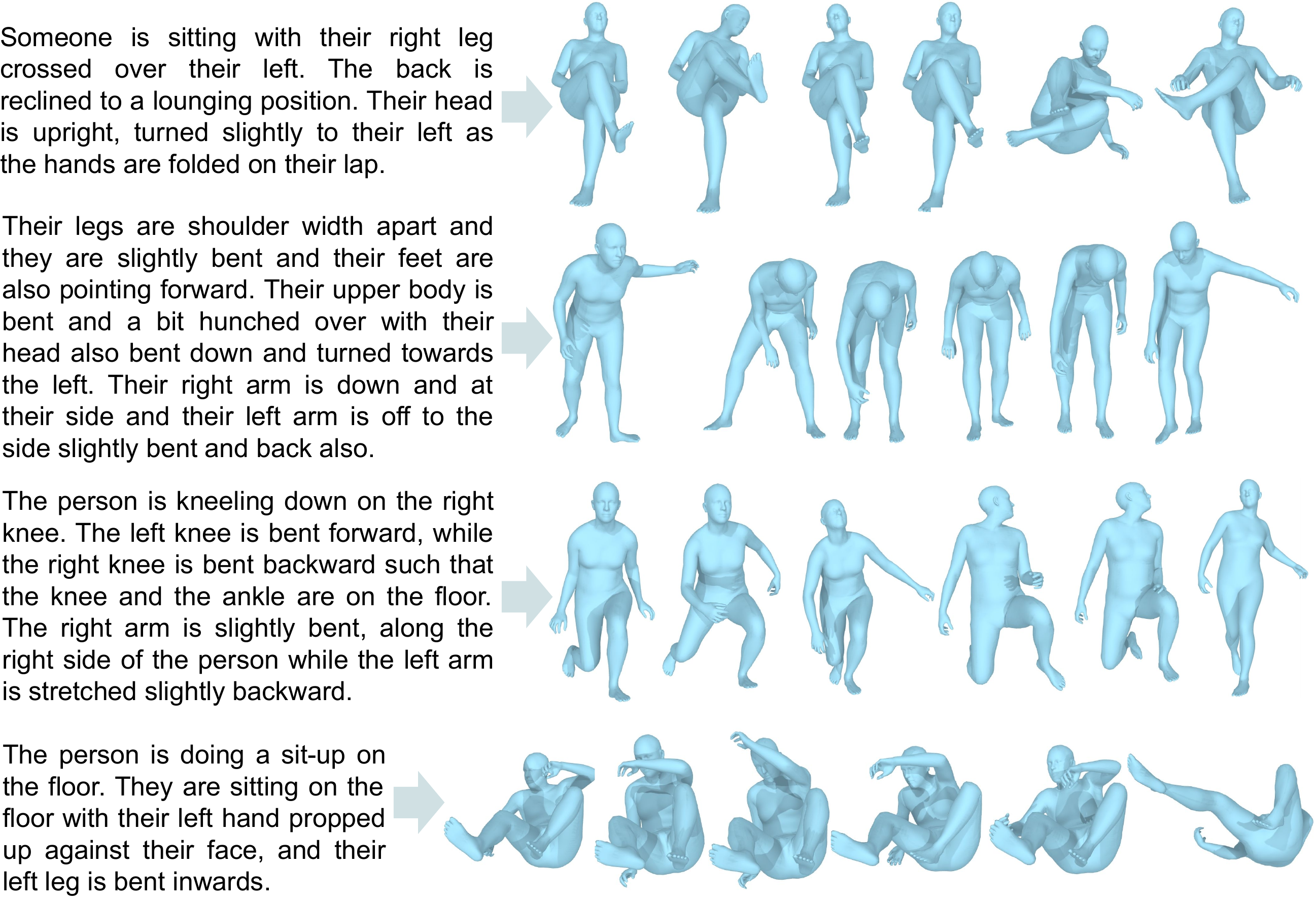} \\[-0.35cm]
    \caption{\textbf{Text-to-pose retrieval results} for human-written captions from  \dname . Directions such as `left' and `right' are relative to the body.}
    \label{fig:t2p_retrieval}
\end{figure}

\myparagraph{Qualitative retrieval results.}
Examples of text-to-pose retrieval results are presented in Figure~\ref{fig:t2p_retrieval}. It appears that the model is able to encode several pose concepts concurrently and to distinguish between the left and right body parts.

\begin{figure*}[t]
    \centering
    \includegraphics[width=0.85\linewidth]{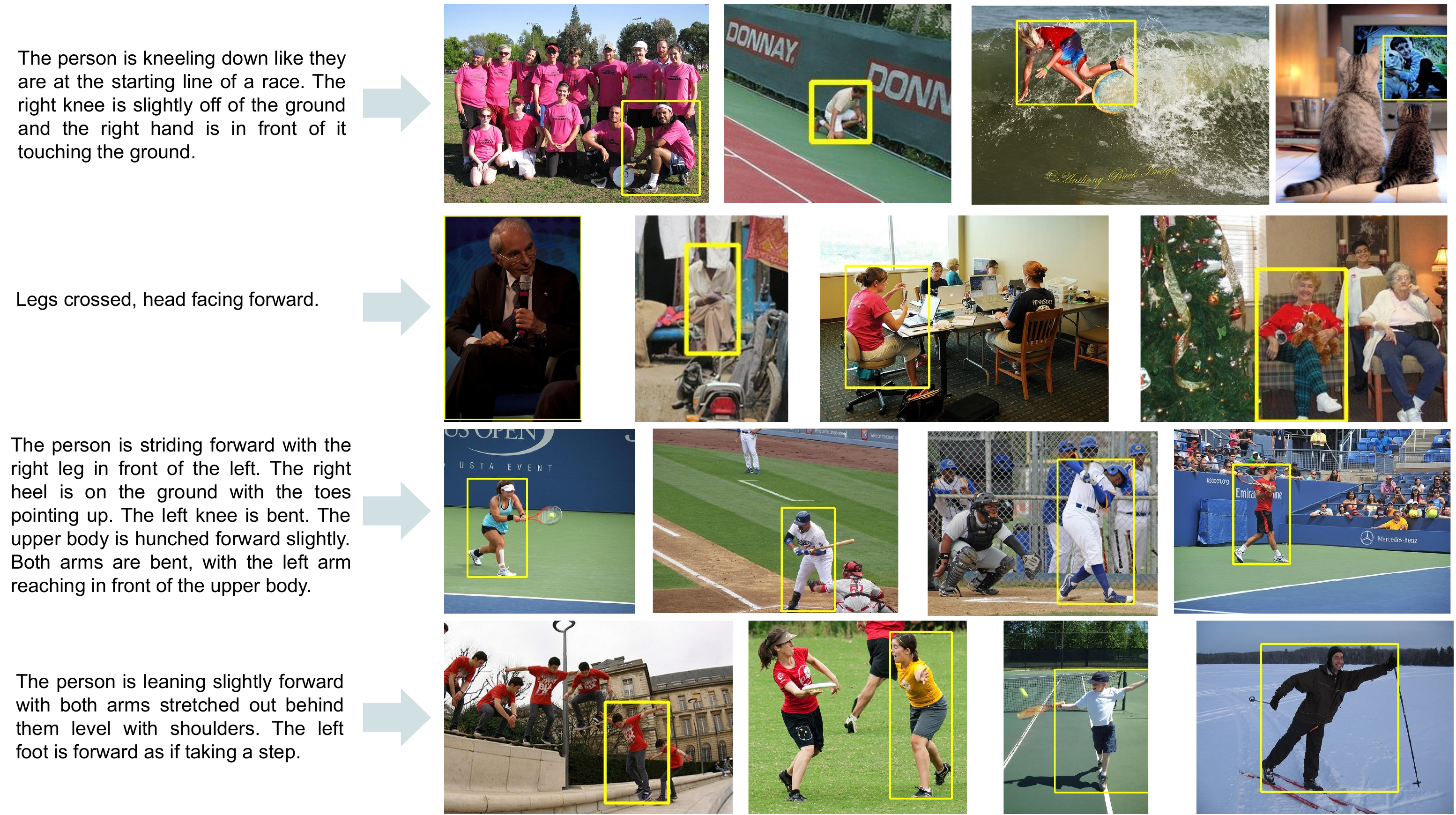} \\[-0.4cm]
    \caption{\textbf{Retrieval results in image databases.} We use our text-to-pose retrieval model trained on human captions from \dname to retrieve 3D poses from SMPL fits on MS Coco, for some given text queries. We display the corresponding pictures and bounding boxes for the top retrieved poses.
    }
    \label{fig:eft_mscoco}
\end{figure*}

\myparagraph{Retrieval in image databases.}
MS Coco~\cite{coco} is one of several real-world datasets that have been used for human mesh recovery. We resort to the 74,834 pseudo-ground-truth SMPL fits provided by EFT~\cite{EFT}, on which we apply our text-to-pose retrieval model trained with \dname. We then retrieve 3D poses among this MS Coco-EFT set, and display the corresponding images with the associated bounding box around the human body. Results are shown in Figure~\ref{fig:eft_mscoco}. We observe that overall, the constraints specified in the query text are satisfied in the images. 
Retrieval is based on the poses and not on the context, hence the third image of the first row where the pose is close to an actual kneeling one.
This shows one application of a retrieval model trained on the PoseScript dataset and applied to a third modality: specific pose retrieval in images. Our model can be applied to any dataset of images containing humans, as long as SMPL fits are also available.

\section{Application to Text-Conditioned Pose Generation}
\label{sec:generation}

\newcommand{\lkld}{\mathcal{L}_{KL}}
\newcommand{\lrecon}{\mathcal{L}_{R}}
\newcommand{\ndist}{\mathcal{N}}

We next study the problem of \emph{text-conditioned human pose generation}, \ie, generating possible matching poses for a given text query. Our proposed model is based on Variational Auto-Encoders (VAEs)~\cite{vae}.

\begin{figure}[t]
    \centering
    \includegraphics[width=\linewidth]{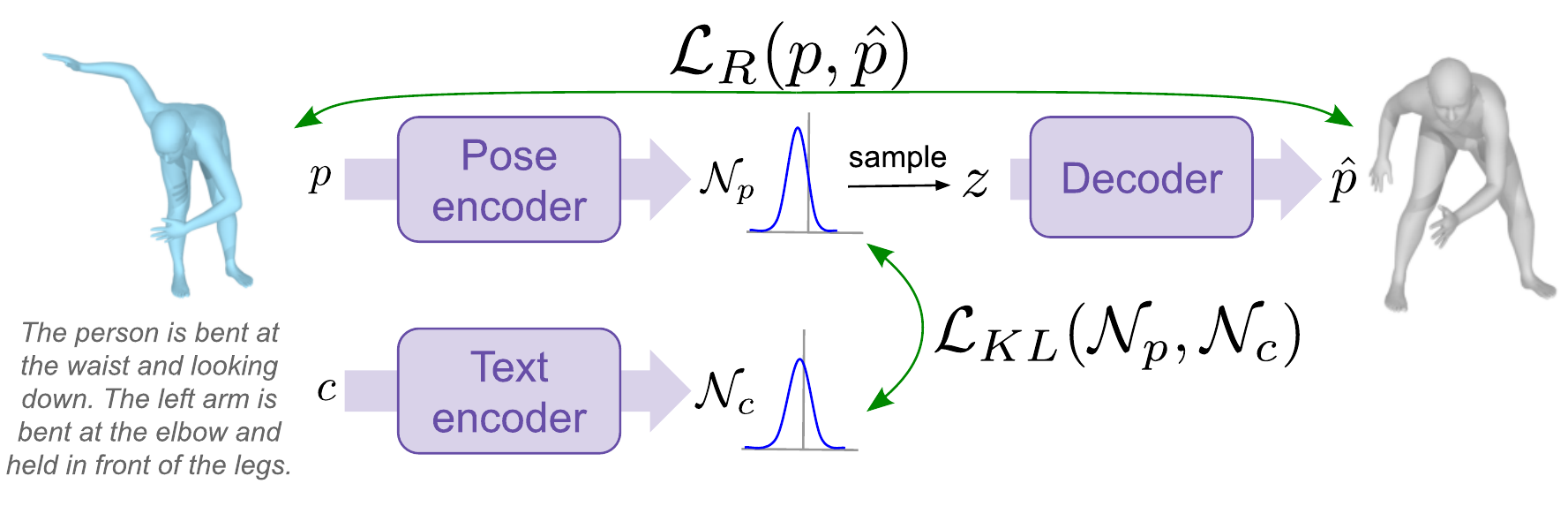} \\[-0.5cm]
    \caption{\textbf{Overview of the text-conditioned generative model.} During training, it follows a VAE but where the latent distribution $\ndist_p$ from the pose encoder has a KL divergence term with the prior distribution $\ndist_c$ given by the text encoder. At test time, the sample $z$ is drawn from $\ndist_c$.}
    \label{fig:generative_model}
\end{figure}

\noindent \textbf{Training.}
Our goal is to generate a pose $\hat{p}$ given its caption $c$. 
To this end, we train a conditional VAE model that takes a tuple $(p,c)$ composed of a pose $p$ and its caption $c$ at training time. Figure~\ref{fig:generative_model} gives an overview of our model. 
A pose encoder maps the pose $p$ to a posterior over latent variables by producing the mean $\mu(p)$ and variance $\Sigma(p)$ of a normal distribution $\ndist_p = \ndist(\cdot | \mu(p), \Sigma(p))$.
Another encoder is used to obtain a prior distribution $\ndist_c$, independent of $p$ but conditioned on $c$. 
A latent variable $z \sim \ndist_p$ is sampled from $\ndist_p$ and decoded into a reconstructed pose $\hat{p}$. The training loss combines a reconstruction term $\lrecon(p,\hat{p})$ between the original and reconstructed poses, $p$ and $\hat{p}$, and a regularization term, the Kullback-Leibler (KL) divergence between $\ndist_p$ and $\ndist_c$:

\begin{equation}
    \mathcal{L} = \lrecon(p,\hat{p}) + \lkld(\ndist_p, \ndist_c).
\end{equation}

We also experiment with an additional loss term $\mathcal{L}_{reg}$ denoting $\lkld(\ndist_p,\ndist(\cdot|0,I))+\lkld(\ndist_c,\ndist(\cdot|0,I))$: KL divergences between the posterior (resp. the prior) and the standard Gaussian $\ndist_0 = \ndist(\cdot|0,I)$. These can be seen as other regularizers and they also allow to sample poses from the model without conditioning on captions.
We treat the variance of the decoder as a learned constant~\cite{sigmavae} and use a negative log likelihood (nll) as reconstruction loss, either from a Gaussian -- which corresponds to an L2 loss and a learned variance term -- or a Laplacian density, which corresponds to an L1 loss. 
Following VPoser, we use SMPL(-H) inputs with the axis-angle representation, and output joint rotations with the continuous 6D representation of~\cite{zhou2019continuity}. Our reconstruction loss $\lrecon(p,\hat{p})$ is a sum of the reconstruction losses between the rotation matrices -- evaluated with a Gaussian log-likelihood -- the position of the joints and the position of the vertices, both evaluated with a Laplacian log-likelihood.

\myparagraph{Text-conditioned generation.} At test time, a caption $c$ is encoded into $\ndist_c$, from which $z$ is sampled and decoded into a generated pose $\hat{p}$.

\begin{table*}[t]
    \caption{\textbf{Evaluation of the text-conditioned generative model} on \dname-A for a model without or with $\mathcal{L}_{reg}$ (top) and on \dname-H without or with pretraining on \dname-A (bottom). Unless specified otherwise, models all have the GloVe-biGRU configuration. Results are averaged over 3 runs. The variability of R/G (resp. G/R) mRecall is due to the random selection of a generated pose sample at test (resp. training) time. For comparison, the mRecall when training and testing on real poses is 72.8 with \dname-A and 45.3 on \dname-H.}
        \centering
        \resizebox{0.8\textwidth}{!}{
        \begin{tabular}{l@{~~}c@{~~}c@{~~}c@{~~}c@{~~}c@{~~}c}
        \toprule
             & \multirow{2}{*}{FID\color{OliveGreen}{$\downarrow$}}
             & ELBO 
             & ELBO  
             & ELBO 
             & mRecall 
             & mRecall   \\ 
             &
             & jts\color{OliveGreen}{$\uparrow$} 
             & vert.\color{OliveGreen}{$\uparrow$}
             & rot.\color{OliveGreen}{$\uparrow$} 
             & R/G\color{OliveGreen}{$\uparrow$} 
             & G/R\color{OliveGreen}{$\uparrow$} \\
        \midrule
        \multicolumn{7}{l}{\textit{test on PoseScript-A}} \\
        ~~~~without $\mathcal{L}_{reg}$ & 0.12 \tiny{${\pm}$ 0.02} & 1.76 \tiny{${\pm}$ 0.00} & 2.05 \tiny{${\pm}$ 0.01} & 1.06 \tiny{${\pm}$ 0.02} & 22.7 \tiny{${\pm}$ 1.4} & \textbf{41.5} \tiny{${\pm}$ 4.5} \\ 
        ~~~~with $\mathcal{L}_{reg}$ &  \textbf{0.07} \tiny{${\pm}$ 0.00} & \textbf{1.78} \tiny{${\pm}$ 0.00} & \textbf{2.09} \tiny{${\pm}$ 0.01} & \textbf{1.10} \tiny{${\pm}$ 0.01} & \textbf{25.1} \tiny{${\pm}$ 0.2} & 40.0 \tiny{${\pm}$ 3.0} \\ 
        \midrule
        \multicolumn{7}{l}{\textit{test on PoseScript-H, for the model with $\mathcal{L}_{reg}$}} \\
        ~~~~without pretraining & 0.29 \tiny{${\pm}$ 0.04} & 0.98 \tiny{${\pm}$ 0.02} & 1.41 \tiny{${\pm}$ 0.02} & 0.53 \tiny{${\pm}$ 0.02} & 5.2 \tiny{${\pm}$ 0.5} & 20.5 \tiny{${\pm}$ 3.4} \\ 
        ~~~~with pretraining & 0.04 \tiny{${\pm}$ 0.01} & 1.39 \tiny{${\pm}$ 0.01} & \textbf{1.75} \tiny{${\pm}$ 0.01} & \textbf{0.87} \tiny{${\pm}$ 0.01} & 19.5 \tiny{${\pm}$ 2.6} & 35.1 \tiny{${\pm}$ 2.5} \\ 
        ~~~~ ~~~~\textit{using the transformer text encoder} & \textbf{0.03} \tiny{${\pm}$ 0.00} & 1.40 \tiny{${\pm}$ 0.00} & 1.74 \tiny{${\pm}$ 0.01} & 0.85 \tiny{${\pm}$ 0.02} & 35.6 \tiny{${\pm}$ 1.2} & 44.5 \tiny{${\pm}$ 0.1} \\ 
        ~~~~ ~~~~ ~~~~\textit{with mirroring augmentation} & \textbf{0.03} \tiny{${\pm}$ 0.00} & \textbf{1.41} \tiny{${\pm}$ 0.00} & \textbf{1.75} \tiny{${\pm}$ 0.01} & 0.86 \tiny{${\pm}$ 0.01} & \textbf{37.5} \tiny{${\pm}$ 0.5} & \textbf{45.2} \tiny{${\pm}$ 1.1} \\ 
        \bottomrule
        \end{tabular}
        }
        \label{tab:gen_auto}
    \end{table*}


\myparagraph{Evaluation metrics.}
We evaluate sample quality following the principle of the Fr\'echet inception distance:
we compare the distributions of features extracted using our retrieval model (see Section~\ref{sec:retrieval}), using real test poses and poses generated from test captions. This is denoted FID with an abuse of notation.
We also report the mean-recall of retrieval models trained on real poses and evaluated on generated poses (mR R/G), and vice-versa (mR G/R). Both metrics are sensitive to sample quality: the retrieval model will fail if the data is unrealistic. The second metric is also sensitive to diversity: missing parts of the data distribution hinder the retrieval model trained on samples.
Finally, we report the Evidence Lower Bound (ELBO) computed on joints, vertices or rotation matrices, normalized by the target dimension.

\begin{figure}[t]
    \centering
    \includegraphics[width=\linewidth]{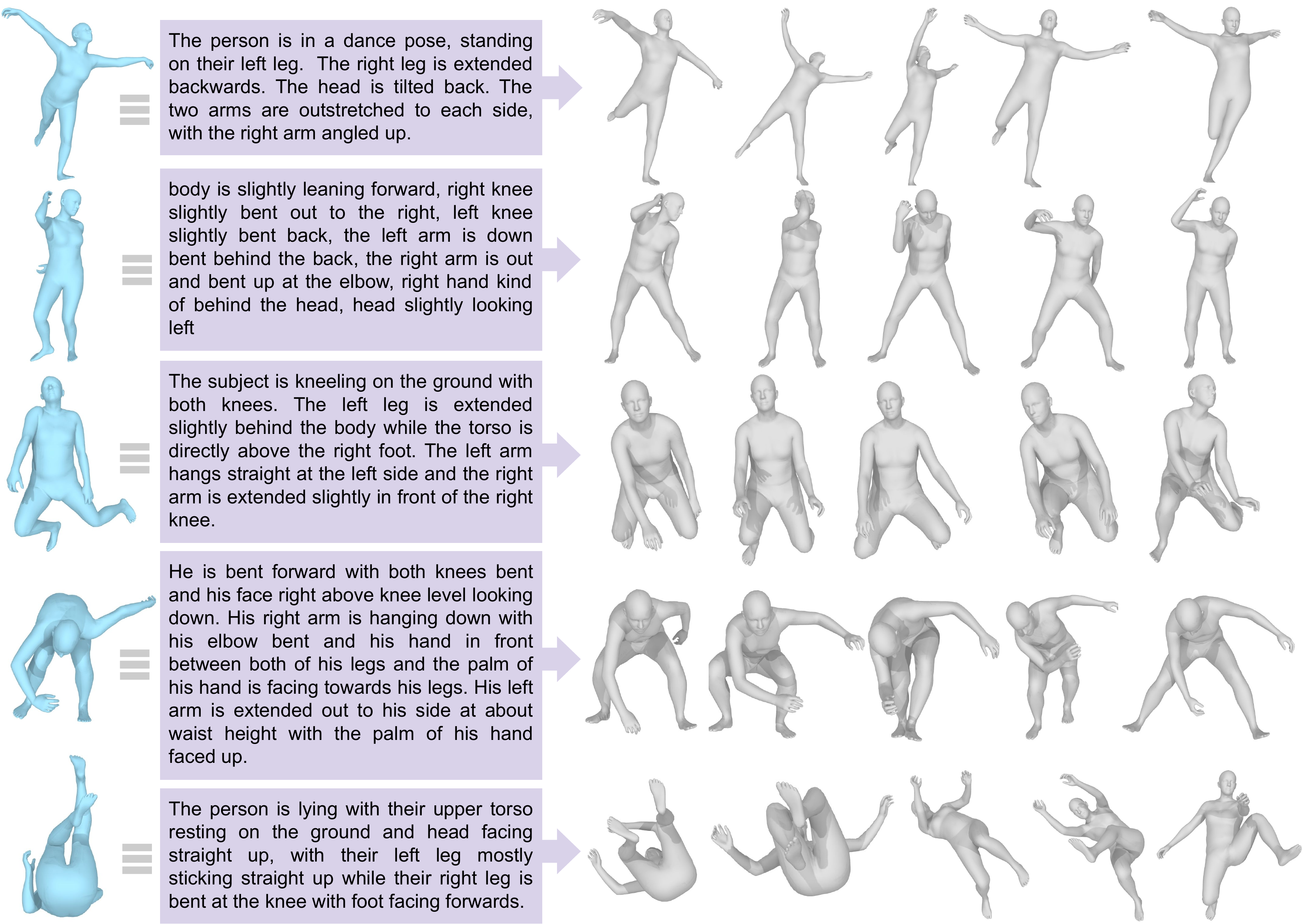} \\[-0.35cm]
    \caption{\textbf{Examples of generated samples.} We show several generated samples (in grey) obtained for the human-written captions presented in the middle. For reference, we also show in blue the pose for which this annotation was originally collected.}
    \label{fig:gen_samples}
\end{figure}

\begin{figure}[t]
    \centering
    \includegraphics[width=0.85\linewidth]{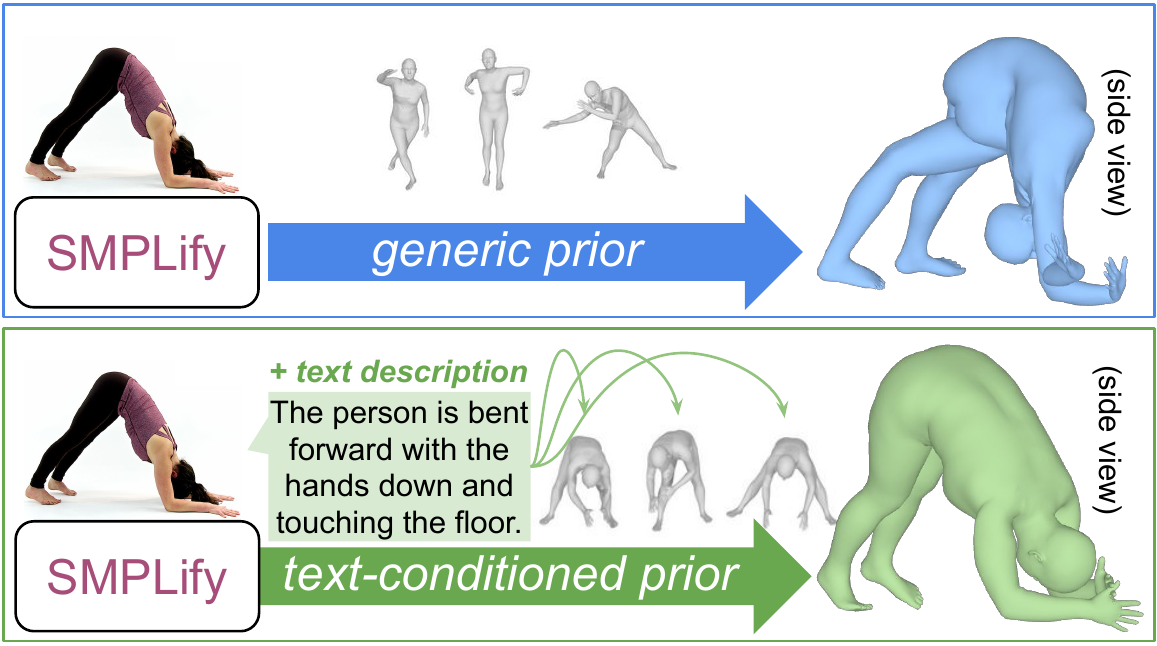} \\[-0.3cm]
    \caption{\textbf{Example of potential application to SMPL fitting in images.} Using the text-conditional pose prior (bottom) yields a more accurate 3D pose than a generic pose prior (top) when running the optimization-based SMPL fitting method SMPLify.}
    \label{fig:smplify_application}
\end{figure}

\myparagraph{Results.}
We present quantitative results in Table~\ref{tab:gen_auto}.
We first find that adding the extra-regularization loss $\mathcal{L}_{reg}$ to the model trained and evaluated on automatic captions is slightly helpful. Also, it is convenient to sample poses without any conditioning. We keep this configuration and evaluate it when (a) training on human captions and (b) pretraining on automatic captions then finetuning on human captions.
Pretraining improves greatly all metrics, showing that it helps to yield realistic and diverse samples. The transformer-based text encoder improves over the one based on GloVe and biGRU, achieving substantially better performance in terms of mRecall. This suggests that the transformer-based text encoder has a finer understanding of the pose semantics.
Slight improvement is yielded by the mirroring augmentation.
We display generated samples in Figure~\ref{fig:gen_samples}; the poses are realistic and generally correspond to the query. There are some variations, especially when the text allows it, for instance with the height of the right leg in the top example or the distance between the legs in the fourth row.
Failure cases can happen; in particular rare words like `lying' in the bottom row lead to higher variance in the generated samples; some of them are nevertheless close to the reference.

\myparagraph{Application to SMPL fitting in image.}
We showcase the potential of leveraging text data for 3D tasks on a challenging example from SMPLify \cite{smplify}, in Figure~\ref{fig:smplify_application}. We use our text-conditional prior instead of the generic VPoser prior \cite{smplx} to initialize to a pose closer to the ground truth and to better guide the in-the-loop optimization. This helps to avoid bad local minima traps.

\section{Application to Pose Description Generation}
\label{sec:text_generation}

We now introduce our \textit{learned} approach to generate pose descriptions in natural language. Conversely to the process presented in Section~\ref{sub:automatic}, this one does not rely on heuristics nor template structures. It is trained on the human-written captions, resulting in generated texts that are more concise, with improved formulation and more high-level concepts. Note that this model does not debase the pipeline from Section~\ref{sub:automatic}: just like the others, it benefits tremendously from the pretraining on the automatic captions the pipeline produces.

\begin{figure}[t]
    \centering
    \includegraphics[width=0.95\linewidth]{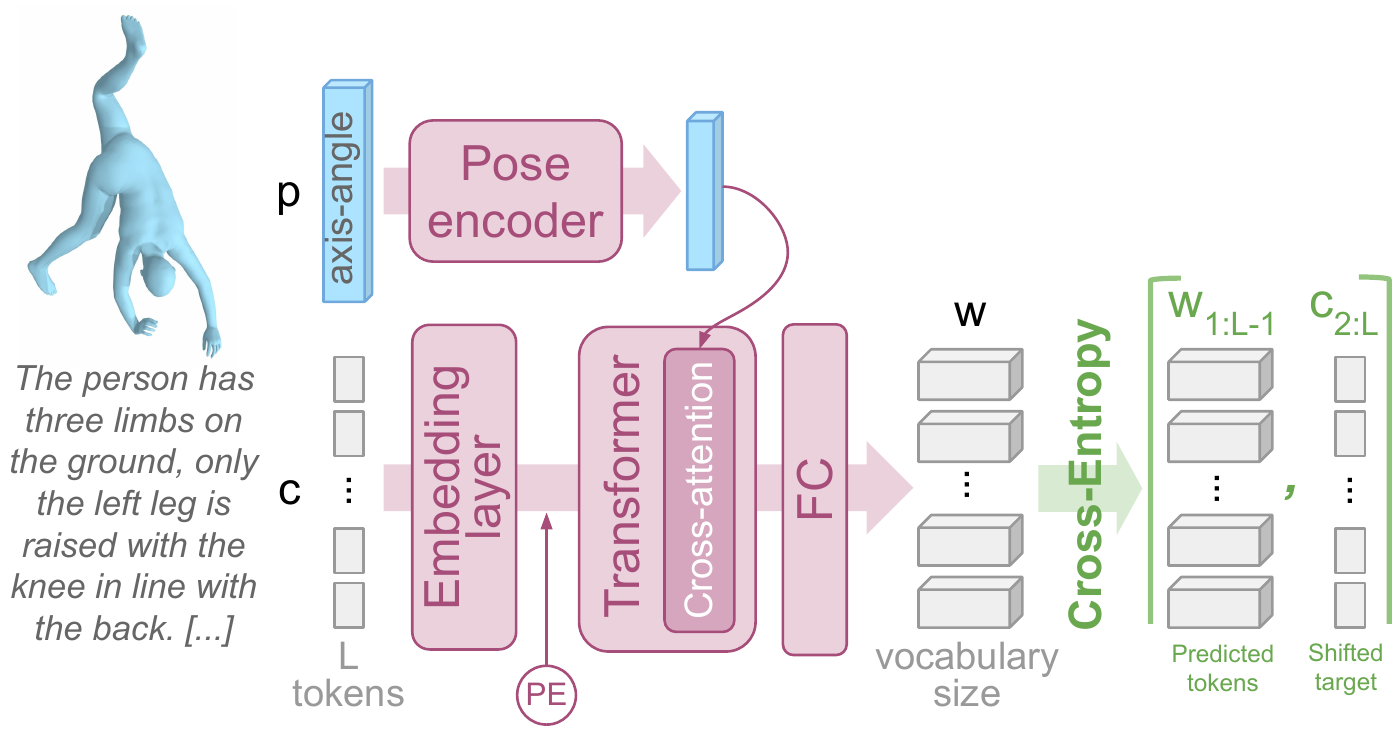} \\[-0.3cm]
    \caption{\textbf{Overview of the pose description generative model.} Pose information is injected in the text transformer via cross-attention. The cross-entropy loss between the output probability distribution over the vocabulary and the shifted target trains the model to recursively predict the next word starting from the \textit{BOS} token.}
    \label{fig:text_generative_model}
\end{figure}

We use an auto-regressive model, that starts from the \textit{BOS} (beginning-of-sequence) token and produces iteratively each new token given all those previously generated, and the pose conditioning; see Figure~\ref{fig:text_generative_model}.

\myparagraph{Training.} The model is provided with the tokenized caption $c_{1:L}$ of $L$ tokens. It embeds each token, adds positional encoding then feeds the result to a transformer, which accounts for the pose conditioning thanks to the cross-attention mechanisms. The use of a causal attention mask prevents the model from attending tokens from $l+1$ to $L$ when dealing with token $l$. Finally, the model outputs a probability distribution $w_{1:L}$ over the vocabulary, where $w_{l}$ corresponds to the probability $p(.|c_{1:l})$. The cross-entropy loss between $w_{1:L-1}$ and $c_{2:L}$ maximizes $p(c_{l+1}|c_{1:l})$, thus training the model to predict the next token $l+1$ from previous ones $c_{1:l}$.

\myparagraph{Inference.} Given the \textit{BOS} token and the input pose, the description is decoded iteratively in a greedy fashion by maximizing the likelihood at each step $l$, until the special token \textit{EOS} is decoded. At step $l$, token $l$ is decoded as the one that maximizes the output $w_l$; it is then appended to the tokens $1$ to $l-1$ decoded earlier, so as to predict token $l+1$ in the next pass.

\myparagraph{Evaluation.} We report the standard natural language processing (NLP) metrics BLEU-4~\cite{Papineni02bleu}, Rouge-L~\cite{lin2004rouge} and METEOR~\cite{banarjee2005meteor}. Following TM2T~\cite{chuan2022tm2t}, we use our retrieval model from Section~\ref{sec:retrieval} to measure the recall at $K\in\{1,2,3\}$ (top-k R-precision) when ranking, for a query pose, the corresponding text generated by our model among 31 randomly sampled generated texts yielded for other poses. Additionally, we report reconstruction metrics (MPJE, MPVE and geodesic distance on the joint rotations) obtained by comparing the input pose with the one generated by our model from Section~\ref{sec:generation}, when given the generated text. While the NLP metrics measure the common n-grams between the reference text and the generated one, the other metrics evaluate the semantic content of the generated description. Indeed, an insufficiently detailed text could not help generate nor retrieve back the input pose.

\myparagraph{Results.} Again, we notice from Table~\ref{tab:text_generative_table} that pretraining on the automatic captions leads to substantially better descriptions, and that the mirroring augmentation helps. We note that R-Precision and reconstruction metrics rely on trained models, and their biased understanding of the data (\ie, some concepts may be ill-encoded). This could explain why the produced texts appear to yield better results than the original ones.

\begin{figure}[t]
    \centering
    \includegraphics[width=\linewidth]{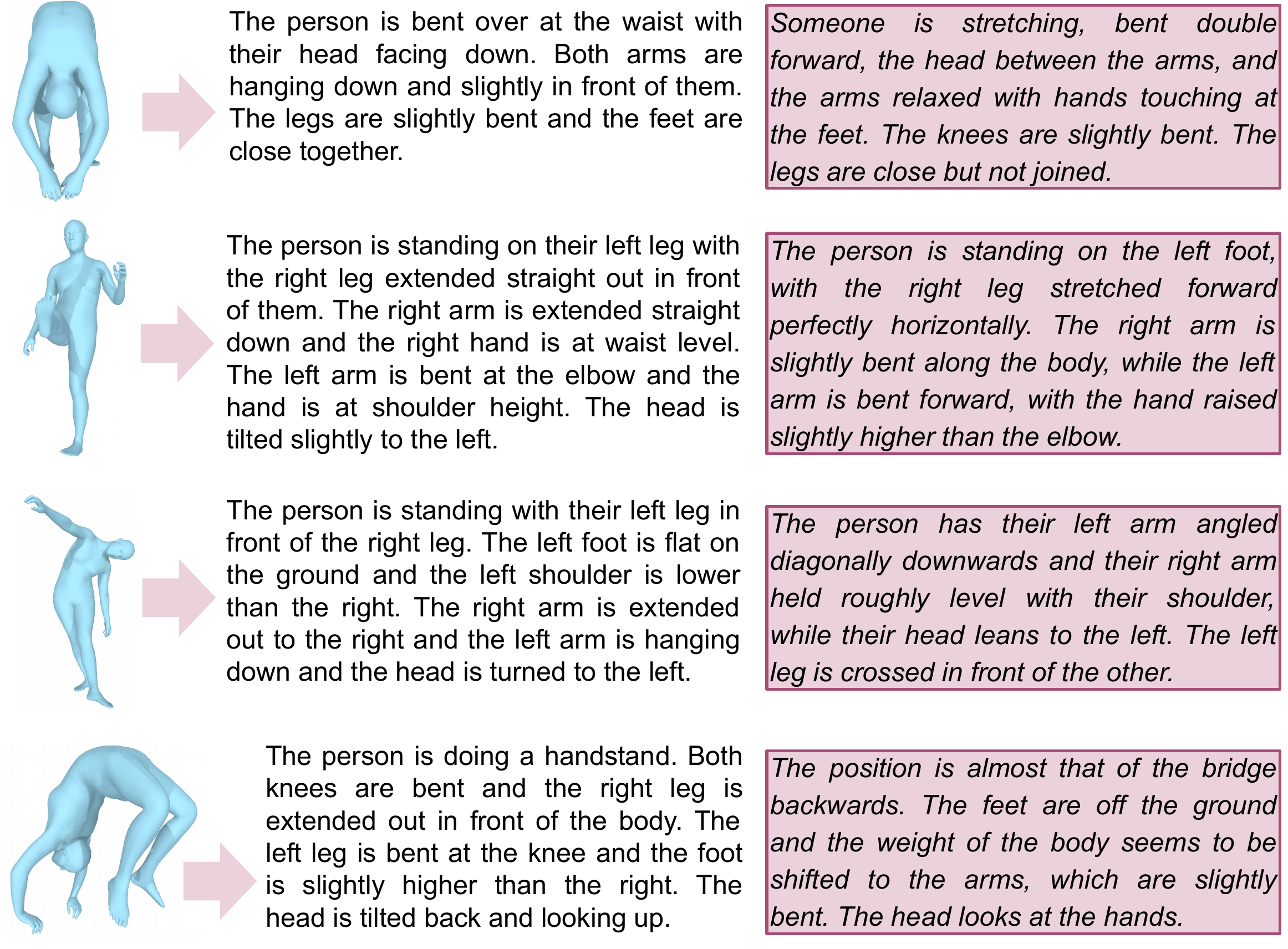} \\[-0.35cm]
    \caption{\textbf{Examples of model-generated descriptions} for the poses on the left. For reference, the annotated texts are presented in the colored boxes.}
    \label{fig:text_generative_qualitatives}
\end{figure}

Examples of descriptions generated by our model are presented in Figure~\ref{fig:text_generative_qualitatives}. It appears that the model is able to produce meaningful descriptions, with egocentric relations and high-level concepts (\eg handstand). However, it sometimes hallucinate (leg position in the last example), or it struggles to understand the pose as a whole, especially in rare cases like upside-down poses (the head would indeed be looking up if the body was not bent backwards that much).

\begin{table*}[t]
    \caption{\textbf{Caption generation results.} The top block shows some reference measures while the lower block evaluates the generated texts for PoseScript-H. Results are averaged over 3 runs.}
    \centering
    \resizebox{\textwidth}{!}{%
    \begin{tabular}{lcccccccccc}
        \toprule
        \multirow{2}{*}{} & \multicolumn{3}{c}{R-Precision $\uparrow$} & \multicolumn{3}{c}{NLP $\uparrow$} & \multicolumn{3}{c}{Reconstruction $\downarrow$ \textit{(best of 30)}} \\
        \cmidrule(lr){2-4} \cmidrule(lr){5-7} \cmidrule(lr){8-10}
        & R@1 & R@2 & R@3 & BLEU-4 & ROUGE-L & METEOR & MPJE & MPVE & Geodesic \\
        \midrule
        random text from PoseScript-H & 3.00 & 5.83 & 8.18 & 7.07 & 23.98 & 25.96 & 419 & 328 & 12.51 \\
        matching text from PoseScript-A & 44.49 & 57.62 & 66.13 & 24.52 & 35.15 & 43.20 & 203 & 170 & 9.13 \\
        matching text from PoseScript-H & 81.60 & 87.60 & 91.25 & 100.00 & 100.00 & 100.00 & 203 & 168 & 8.93 \\
        \midrule
        without pretraining & 24.39 \tiny{${\pm}$ 2.34} & 36.01 \tiny{${\pm}$ 2.22} & 43.87 \tiny{${\pm}$ 1.78} & 11.35 \tiny{${\pm}$ 0.13} & 31.32 \tiny{${\pm}$ 0.22} & 30.88 \tiny{${\pm}$ 0.19} & 324 \tiny{${\pm}$ 1} & 254 \tiny{${\pm}$ 2} & 10.74 \tiny{${\pm}$ 0.04} \\
        with pretraining & 88.09 \tiny{${\pm}$ 0.64} & 93.33 \tiny{${\pm}$ 0.33} & 95.54 \tiny{${\pm}$ 0.07} & 13.07 \tiny{${\pm}$ 0.11} & 33.86 \tiny{${\pm}$ 0.16} & 32.95 \tiny{${\pm}$ 0.20} & 203 \tiny{${\pm}$ 1} & 169 \tiny{${\pm}$ 0} & 8.80 \tiny{${\pm}$ 0.02} \\
        ~~~~ \textit{with mirroring augmentation} & \textbf{88.98} \tiny{${\pm}$ 1.51} & \textbf{94.46} \tiny{${\pm}$ 0.53} & \textbf{96.14} \tiny{${\pm}$ 0.34} & \textbf{13.22} \tiny{${\pm}$ 0.17} & \textbf{34.07} \tiny{${\pm}$ 0.04} & \textbf{33.09} \tiny{${\pm}$ 0.21} & \textbf{200} \tiny{${\pm}$ 2} & \textbf{166} \tiny{${\pm}$ 2} & \textbf{8.75} \tiny{${\pm}$ 0.07} \\
        \bottomrule
    \end{tabular}
    \label{tab:text_generative_table}
    }
\end{table*}

\section{Characteristics of automatic captions}
\label{sec:caracteristics_auto_cap}

In this section, we aim to study the impact of different aspects of the automatic captioning pipeline. To this end, we generate 6 different kinds of captions per pose, each with different characteristics: we generate all the captions using the same pipeline, which is presented in Section~\ref{sec:dataset}, and disable some steps of the process to produce the different versions.

Specifically, steps that were deactivated include: (1) randomly skipping eligible \posecodes for description; (2) aggregating \posecodes (``implicitness''), omitting support keypoints (\eg `the right foot is behind the torso' does not turn into `the right foot is in the back' when this step is deactivated) and randomly referring to a body part by a substitute word (\eg `it'/`they', `the other'); (3) adding a sentence constructed from high-level pose annotations given by BABEL~\cite{babel}; and (4) removing redundant \posecodes based on ripple effect rules.

Among all 100k poses of \dname, only 36,317 are annotated in BABEL and may benefit from an additional sentence in their automatic description.
As 28\% of \dname poses come from DanceDB, which was not annotated in BABEL, we additionally assign the `dancing' label to those DanceDB-originated poses, for one variant of the automatic captions that already leverages BABEL auxiliary annotations. This results in 64,758 poses benefiting from an auxiliary label.

Table~\ref{tab:cap_version_summary} summarises the characteristics of the 6 captions versions introduced in this section ($N1$ to $N5$), along with those of the captions versions composing \dname-A (last row), which is used in the rest of this article.


\begin{table}
    \caption{\textbf{Summary of the automatic caption versions.} \tx~ symbols indicate when characteristics apply to each caption version. All models were trained on a pool of 3 captions per pose (multiplicity). Mean recall results are averaged over 3 runs of models trained with the bi-GRU configuration.}
    \centering
    \resizebox{\linewidth}{!}{
    \begin{tabular}{cccccc|c}
    \toprule
    \begin{tabular}{c}Version\\(pretraining)\end{tabular} & Multiplicity & \begin{tabular}{c}Random\\skip\end{tabular} & Implicitness & Auxiliary labels & \begin{tabular}{c}Ripple\\effect\end{tabular} & mRecall \\
    \midrule
    \texttt{N1} & $\times 3$ & - & - & - & - & 37.5 \tiny{${\pm}$ 1.6} \\ 
    \texttt{N2} & $\times 3$ & \tx & - & - & - & 38.8 \tiny{${\pm}$ 1.3} \\ 
    \texttt{N3} & $\times 3$ & - & \tx & - & - & 39.8 \tiny{${\pm}$ 0.5} \\ 
    \texttt{N4} & $\times 3$ & - & - & {\scriptsize (w/o dancing label)} & - &  37.6 \tiny{${\pm}$ 0.7} \\ 
    \texttt{N4d} & $\times 3$ & - & - & {\scriptsize (w/ dancing label)} & - & 38.7 \tiny{${\pm}$ 1.9} \\ 
    \texttt{N5} & $\times 3$ & - & - & - & \tx & 37.1 \tiny{${\pm}$ 2.0} \\ 
    \midrule
    \multirow{3}{*}{PoseScript-A} & $\times 1$ & \tx & - & - & - & \multirow{3}{*}{40.9 \tiny{${\pm}$ 0.1}} \\
                                  & $\times 1$ & \tx & \tx & {\scriptsize (w/ dancing label)} & - & \\ 
                                  & $\times 1$ & \tx & \tx & - & \tx & \\ 
    \bottomrule
    \end{tabular}}
    \vspace{1mm}
    \label{tab:cap_version_summary}
\end{table}


We pretrain 6 different retrieval models (GloVe+biGRU configuration), one with each caption version $N1$ to $N5$. Each model is trained on a pool of 3 generated captions from the same version. Next, we finetune the retrieval models on \dname-H. We report the mean recall on the test set of \dname-H in the last column of Table~\ref{tab:cap_version_summary}.

Due to relatively high variability, it is hard to tell which aspects of the automatic captioning pipeline are the most important. However, it appears clearly that implicitness (\ie, posecode aggregations) make for automatic captions that are closer to human-written ones, as leveraging captions $N3$ for pretraining achieve the best results. Eventually, the best performance is obtained by pretraining the model on a pool of 3 captions from different versions (last row). This last pool of captions, namely \dname-A, is the one used in the experiments for all other sections.
\section{Size of the training data}
\label{sec:size_training_data}

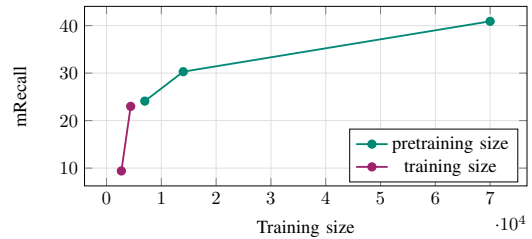
\begin{figure}[t]
    \center
    \resizebox{0.8\linewidth}{!}{
    \begin{tikzpicture}
        \begin{axis}[
        width=10cm,
        height=5cm,
        xlabel=Training size,
        ylabel=mRecall,
        grid=both,
        minor grid style={gray!25},
        major grid style={gray!25},
        legend style={at={(0.98,0.02)},anchor=south east},
        ]
        \addplot[line width=1pt,solid,color=PineGreen,mark=*,restrict x to domain=6000:100000] %
        table[x=datasize,y=mrecall,col sep=comma]{datasize.csv};
        \addlegendentry{pretraining size};
        \addplot[line width=1pt,solid,color=RedViolet,mark=*,restrict x to domain=0:6000] %
        table[x=datasize,y=mrecall,col sep=comma]{datasize.csv};
        \addlegendentry{training size};
    \end{axis}
    \end{tikzpicture}
    }
    \vspace{-3mm}
\caption{\textbf{Training with different amounts of data.} Results are reported on the test split of \dname-H$_{2}$. The \textcolor{PineGreen}{green} curve corresponds to the size of automatic data used for pretraining, while the \textcolor{RedViolet}{violet} curve corresponds to the size of human-written data used for direct training.}
\label{fig:data_size_results}
\end{figure}


In this section, we study the impact of the number of automatic captions used for pretraining. To this end, we define data subsets of different sizes with automatic (10k, 20k and 100k) or human-written annotations (3.9k and 6.3k).
We train one retrieval model (GloVe+biGRU configuration) for each of the \dname-A subsets, and finetune it on \dname-H. Besides, we train retrieval models directly on each human-written data subsets. We compare results on the test set of \dname-H in Figure~\ref{fig:data_size_results}.

We observe better results when leveraging larger amounts of automatic captions at pretraining time, especially when the number of automatic captions exceeds significantly the number of human-written descriptions.
\section{Implementation details}
\label{app:implementation_details}

We follow VPoser \cite{smplx} for the pose encoder and decoder architectures (except that we use the 52 joints of SMPL-H~\cite{mano}). GloVe word embeddings are 300-dimensional. We use a one-layer bidirectional GRU with 512-dimensional hidden state features. Our transformer-based~\cite{transformer} text encoder uses frozen DistilBERT~\cite{sanh2019distilbert} word embeddings of dimension 768, which are then passed to a ReLU and projected into a 512-dimensional space. We next apply a cosine positional encoding, and feed the result to a transformer composed of 4 layers, 4 heads, GELU activations, and feed-forward networks of size 1024. The final embedding of the text sequence is obtained by average-pooling of the output. We use the same encoders for retrieval and generative tasks. Our text decoder is a transformer with the same configuration as for the text encoder, except that it has 8 heads; both the word embeddings and the latent space are of size 512.
Models are optimized with Adam~\cite{kingma2014adam}. We use an initial loss temperature of $\gamma = 10$ for the retrieval model, and a learning rate coefficient of $0.1$ for the pose auto-encoder in the pose generative model at finetuning time. See Table~\ref{tab:implem_details} for details.

\begin{table}[h]
    \caption{\textbf{Implementation details.} $d$ is the pose embedding size, \textit{lr} stands for `learning rate'. The several values are for training on PoseScript-A and PoseScript-H respectively.}
    \vspace{-3mm}
    \centering
    \resizebox{\linewidth}{!}{%
    \begin{tabular}{lcccccccc@{}}
    \toprule
    model & epochs & batch size & $d$ & lr init & lr or weight decay \\
    \midrule
    retrieval & 1000 / 300 & 512 / 32 & 512 & $2 . 10^{-4}$ & 0.5 every 400 / 40 epochs \\
    pose generative & 2000 / 1000 & 128 & 32 & $10^{-4}$ / $10^{-5}$ & $10^{-4}$ \\
    text generative & 3000 / 2000 & 128 & $32 \to 512$ & $10^{-4}$ / $10^{-5}$ & $10^{-4}$ \\
    \bottomrule
    \end{tabular}
    }
    \label{tab:implem_details}
\end{table}
\section{Discussion and conclusion}

We introduced \dname, the first dataset to map 3D human poses and descriptions in natural language. 
We provided multi-modal applications to text-to-pose retrieval, to text-conditioned human pose generation and pose description generation.
Pretraining on the automatic texts improves performance notably (by a factor 2), systematically over all three studied tasks.

\myparagraph{Limitations.} The accuracy of our models depends in high part on the training data. For instance, our models struggle due to the limited amount of self-contact or upside-down poses.
An alternative to collecting more human-written texts would be to design specific posecodes describing such pose configurations. Another general observation is that our models strive to produce results meeting \textit{all} the text requirements, as the PoseScript descriptions are very rich and complex.

\myparagraph{Future works.} The PoseScript dataset could be extended to account for multi-people interactions. One could also leverage knowledge from large multi-modal models (\eg text-to-image) to help filling in the gaps of the collected data in some aspects (\eg activity concepts). One could further explore the use of a text-based pose prior (\ie with body semantics awareness) for other applications, \eg action recognition.

\section*{Acknowledgments}
This work is supported by the Spanish government with the project MoHuCo PID2020-120049RB-I00, and by NAVER LABS Europe under technology transfer contract `Text4Pose'.

\bibliography{biblio}
\bibliographystyle{IEEEtran}


\vspace{-15mm}
\begin{IEEEbiography}
[{\includegraphics[width=1in,height=1.25in,clip,keepaspectratio]{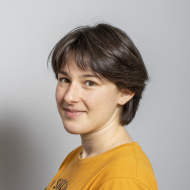}}]{Ginger Delmas} is a PhD student at Institut de Robòtica i Informàtica Industrial (CSIC-UPC) and NAVER LABS Europe since 2021, supervised by Francesc Moreno-Noguer, Philippe Weinzaepfel and Gr\'egory Rogez. She received a M.Sc. degree from Institut Polytechnique de Paris and an Engineering degree in Computer Sciences from Télécom Paris in 2020. Her research is centered on leveraging text for 3D human poses.
\end{IEEEbiography}

\vspace{-18mm}
\begin{IEEEbiography}[{\includegraphics[width=1in,height=1.25in,clip,keepaspectratio]{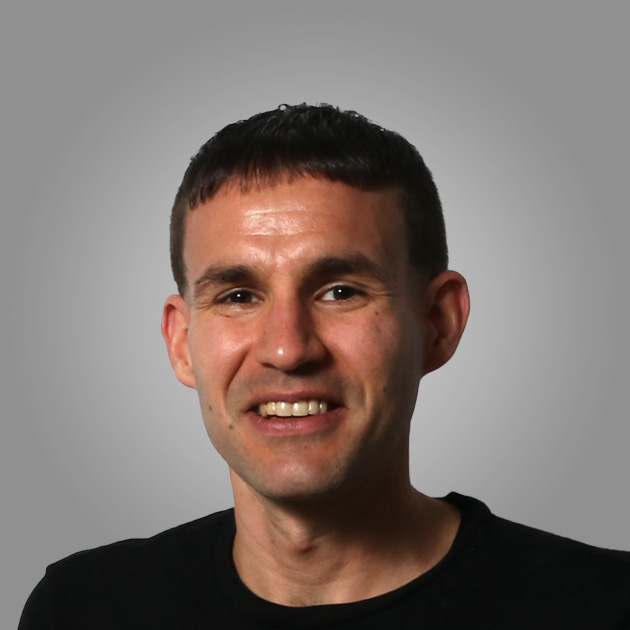}}]{Philippe Weinzaepfel} received a M.Sc. degree from Universite Grenoble Alpes, France, and Ecole Normale Superieure de Cachan, France, in 2012. He was a PhD student in the Thoth team, at Inria Grenoble and LJK, from 2012 until 2016, and received a PhD degree in computer science from Universite Grenoble Alpes in 2016. He is currently a Senior Research Scientist at NAVER LABS Europe, France. His research interests include computer vision and machine learning, with special interest in representation learning and human pose estimation.
\end{IEEEbiography}

\vspace{-15mm}
\begin{IEEEbiography}
[{\includegraphics[width=1in,height=1.25in,clip,keepaspectratio]{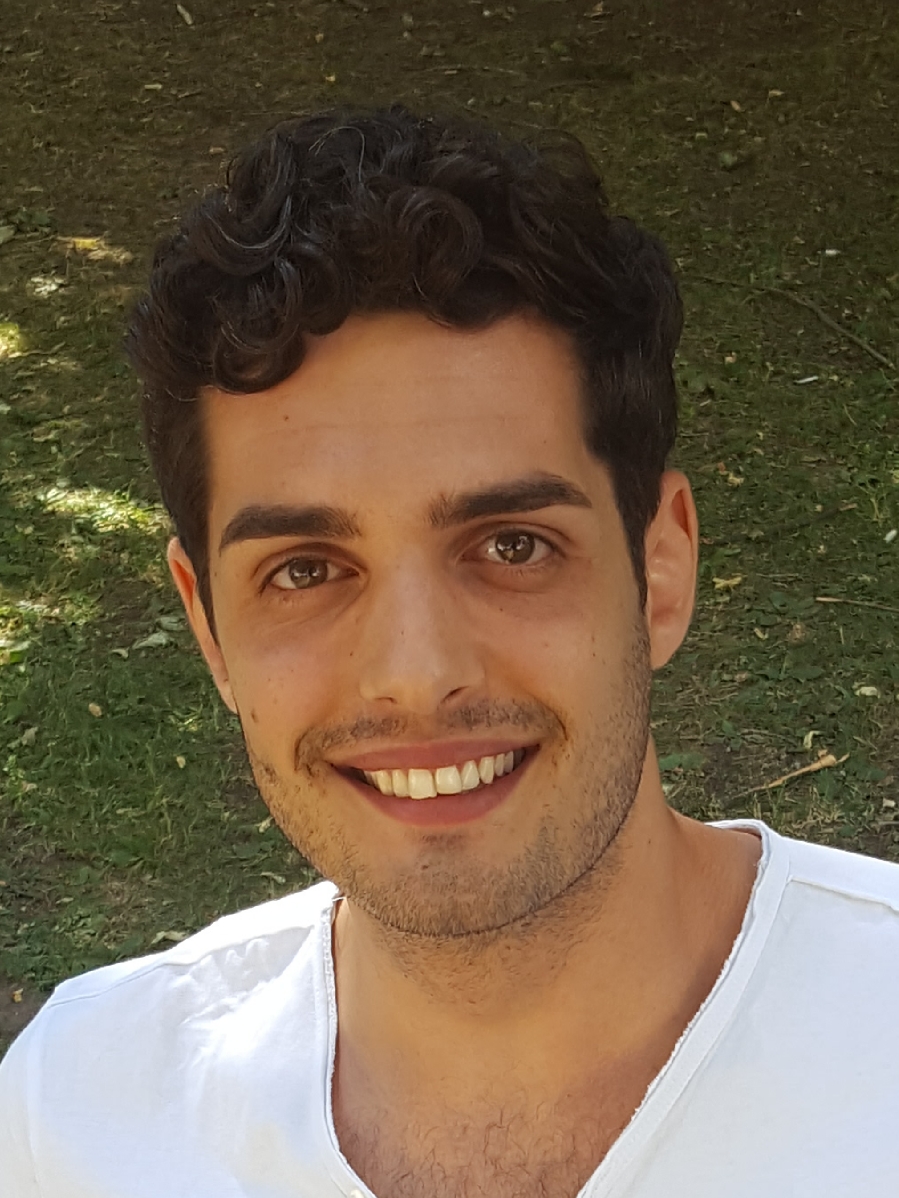}}]{Thomas Lucas} is a Research Scientist at NAVER LABS Europe, since December 2020, in the fields of computer vision and machine learning. He obtained his PhD from INRIA Grenoble, under Jakob Verbeek and Karteek Alahari's supervision. He received an Engineer's degree (MSc) from Ensimag, in applied mathematics and informatics. His research interests revolve around generative modelling of images, with unsupervised or semi-supervised representation learning.
\end{IEEEbiography}

\vspace{-18mm}
\begin{IEEEbiography}
[{\includegraphics[width=1in,height=1.25in,clip,keepaspectratio]{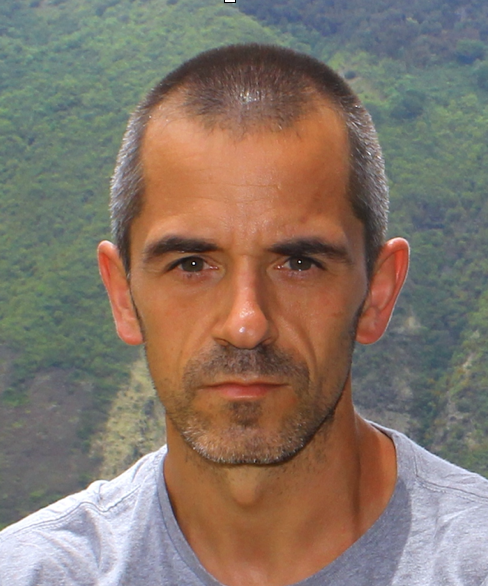}}]{Francesc Moreno-Noguer} is a Research Scientist of the Spanish National Research Council at the Institut de Robotica i Informatica Industrial. His research interests are in Computer Vision and Machine Learning, with topics including human shape and motion estimation, 3D reconstruction of rigid and nonrigid objects and camera calibration. He received the Polytechnic University of Catalonia’s Doctoral Dissertation Extraordinary Award, several best paper awards (e.g. ECCV 2018 Honorable mention, ICCV 2017 workshop in Fashion, Intl. Conf. on Machine Vision applications 2016), outstanding reviewer awards at ECCV 2012 and CVPR 2014, 2021, and Google and Amazon Faculty Research Awards in 2017 and 2019, respectively. He has (co)authored over 150 publications in refereed journals and conferences (including 10 IEEE Transactions on PAMI, 5 Intl. Journal of Computer Vision, 28 CVPR, 11 ECCV and 9 ICCV).
\end{IEEEbiography}

\vspace{-15mm}
\begin{IEEEbiography}[{\includegraphics[width=1in,height=1.25in,clip,keepaspectratio]{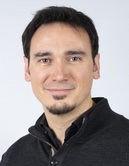}}]{Grégory Rogez} graduated from Centrale Marseille in 2002 and received the M.Sc. degree in biomedical engineering and the Ph.D. degree in computer vision from the University of Zaragoza, Spain, in 2005 and 2012 respectively. His work on monocular human body pose analysis received the best Ph.D. thesis award from the Spanish Association on Pattern Recognition (AERFAI) for the period 2011-2013. He was a regular visiting Research Fellow at Oxford Brookes University (2007-2010), a Marie Curie Fellow at the University of California, Irvine (2013-2015),  a Research Scientist with the LEAR/THOTH team at Inria Grenoble Rhône-Alpes (2015-2018) and since 2019 he is a Senior Research Scientist at NAVER LABS Europe where he leads the computer vision group. His research interests include computer vision and machine learning, with a special focus on understanding people from visual data. 
\end{IEEEbiography}

\clearpage
\section*{Supplementary Material}
\appendices
In this supplementary material, we give additional details on how to compute the different kinds of \posecodes in Section~\ref{app:posecodes}, and specify a list of those that are used in our work.
In Section~\ref{app:auto_captioning_details}, we provide additional information about our automatic captioning pipeline. Other miscellaneous discussions can be found in Section~\ref{app:misc_discussions}.

\section{\Posecodes}
\label{app:posecodes}

\subsection{Computing \posecodes}

We detail here how the different kinds of \posecodes are computed.

\myparagraph{Elementary \posecodes.}

\vspace{0.1cm}
\itemACP{Angle \posecodes} describe how a body part `bends' around a joint $j$. Let a set of keypoints $(i,j,k)$ where $i$ and $k$ are neighboring keypoints to $j$ -- for instance left shoulder, elbow and wrist respectively -- and let $p_l$ denote the position of keypoint $l$. The angle \posecode is computed as the cosine similarity between vectors $v_{ji} = p_i - p_j$ and $v_{jk} = p_k - p_j$.

\vspace{0.1cm}
\itemACP{Distance \posecodes} rate the $L2$-distance $\Vert v_{ij} \Vert$ between two keypoints $i$ and $j$.

\vspace{0.1cm}
\itemACP{\Posecodes on relative position} compute the difference between two sets of coordinates along a specific axis, to determine their relative positioning. A keypoint $i$ is `\texttt{at the left of}' another keypoint $j$ if $p_i^x > p_j^x$; it is `\texttt{above}' it if $p_i^y > p_j^y$;  and `\texttt{in front of}' it if $p_i^z > p_j^z$.

\vspace{0.1cm}
\itemACP{Pitch \& roll \posecodes} assess the verticality or horizontality of a body part defined by two keypoints $i$ and $j$. A body part is said to be `\texttt{vertical}' if the cosine similarity between $\frac{v_{ij}}{\Vert v_{ij} \Vert}$ and the unit vector along the $y$-axis is close to 0. A body part is said to be `\texttt{horizontal}' if it is close to 1.

\vspace{0.1cm}
\itemACP{Ground-contact \posecodes} can be seen as specific cases of relative positioning \posecodes along the $y$ axis. They help determine whether a keypoint $i$ is close to the ground by evaluating
\(\displaystyle p_i^y - \text{min}_{j} p_j^y\).
As not all poses are semantically in actual contact with the ground, we do not resort to these \posecodes for systematic description, but solely for intermediate computations, to further infer super-\posecodes for specific pose configurations.

\myparagraph{Randomized binning step.} As described above, each type of \posecode is first associated to a value $v$ (a cosine similarity angle or a distance), then binned into categories using predefined thresholds.
In practice, hard deterministic thresholding is unrealistic as two different persons are unlikely to always have the same interpretation when the values are close to category thresholds, \eg when making the distinction between `\texttt{spread}' and `\texttt{wide}'.
Thus the categories are inherently ambiguous and to account for this human subjectivity, we randomize the binning step by defining a tolerable noise level $\eta_{\tau}$ on each threshold $\tau$. We then categorize the \posecode by comparing $v+\epsilon$ to $\tau$, where $\epsilon$ is randomly sampled in the range $[-\eta_\tau,\eta_\tau]$.
Hence, a given pose configuration does not always yield the exact same \posecode categorization.

\myparagraph{Super-\posecodes } are binary, and are not subject to the binning step. They only apply to a pose if all of the elementary \posecodes they are based on possess  the respective required \posecode categorization.

\subsection{List of \posecodes}

The list of the 77 elementary \posecodes that are used in our work includes 4 angle \posecodes, 22 distance \posecodes, 34 \posecodes describing relative positions (7 along the $x$-axis, 17 along the $y$-axis and 10 along the $z$-axis), 13 pitch \& roll \posecodes and 4 ground-contact \posecodes. We specify the keypoints involved in the computation of each of these \posecodes in Table~\ref{tab:elementary_posecodes}. Conditions for \posecode categorizations (\ie, thresholds applied to the measured angles and distances, with the corresponding random noise level) are indicated for each kind of \posecode in Table~\ref{tab:elementary_posecodes_thresholds}.
Some of these elementary \posecodes can be combined into super-\posecodes. We list the 10 super-\posecodes we currently consider in Table~\ref{tab:super_posecodes}, and indicate for each of them the different ways they can be produced from elementary \posecodes.

\begin{table*}[t]
\caption{\textbf{List of elementary \posecodes.} We provide the keypoints involved in each of the \posecodes, for each type of elementary \posecodes (angle, distance, relative position, pitch \& roll or ground-contact). We grouped \posecodes on relative positions for better readability, as some keypoints are studied along several axes (considered axes are indicated in parenthesis). Letters `L' and `R' stand for `left' and `right' respectively. Ignored, skippable and unskippable \posecodes are shown in Figures~\ref{fig:angle_posecodes}, \ref{fig:distance_posecodes}, \ref{fig:relposX_posecodes}, \ref{fig:relposY_posecodes}, \ref{fig:relposZ_posecodes}, \ref{fig:pitchroll_posecodes} and \ref{fig:groundcontact_posecodes}.}
\centering \scriptsize
\begin{tabular}{l@{\hskip 0.7cm}c@{\hskip 0.7cm}r}
\begin{tabular}{c}
        \toprule
        \textit{Angle \posecodes} \\
        \midrule
        L-knee  \\
        R-knee  \\
        L-elbow \\
        R-elbow \\
        \toprule
        \textit{Distance \posecodes} \\
        \midrule
        L-elbow \vs R-elbow     \\
        L-hand \vs R-hand       \\
        L-knee \vs R-knee       \\
        L-foot \vs R-foot       \\
        L-hand \vs L-shoulder   \\
        L-hand \vs R-shoulder   \\
        R-hand \vs L-shoulder   \\
        R-hand \vs R-shoulder   \\
        L-hand \vs R-elbow      \\
        R-hand \vs L-elbow      \\
        L-hand \vs L-knee       \\
        L-hand \vs R-knee       \\
        R-hand \vs L-knee       \\
        R-hand \vs R-knee       \\
        L-hand \vs L-ankle      \\
        L-hand \vs R-ankle      \\
        R-hand \vs L-ankle      \\
        R-hand \vs R-ankle      \\
        L-hand \vs L-foot       \\
        L-hand \vs R-foot       \\
        R-hand \vs L-foot       \\
        R-hand \vs R-foot       \\
        \bottomrule
    \end{tabular}&
    \begin{tabular}{c}
        \toprule
        \textit{Ground-contact \posecodes} \\
        \midrule
        L-knee                        \\
        R-knee                        \\
        L-foot                        \\
        R-foot                        \\
        \toprule
        \textit{Relative position \posecodes} \\
        \midrule
        L-shoulder \vs R-shoulder (YZ)\\
        L-elbow \vs R-elbow (YZ)      \\
        L-hand \vs R-hand (XYZ)       \\
        L-knee \vs R-knee (YZ)        \\
        R-foot \vs R-foot (XYZ)       \\
        neck \vs pelvis (XZ)           \\
        L-ankle \vs neck (Y)           \\
        R-ankle \vs neck (Y)           \\
        L-hip \vs L-knee (Y)           \\
        R-hip \vs R-knee (Y)           \\
        L-hand \vs L-shoulder (XY)     \\
        R-hand \vs R-shoulder (XY)     \\
        L-foot \vs L-hip (XY)          \\
        R-foot \vs R-hip (XY)          \\
        L-wrist \vs neck (Y)           \\
        R-wrist \vs neck (Y)           \\
        L-hand \vs L-hip (Y)           \\
        R-hand \vs R-hip (Y)           \\
        L-hand \vs torso (Z)           \\
        R-hand \vs torso (Z)           \\
        L-foot \vs torso (Z)           \\
        R-foot \vs torso (Z)           \\
        \bottomrule
    \end{tabular}&
    \begin{tabular}{c}
        \toprule
        \textit{Pitch \& roll \posecodes} \\
        \midrule
        L-hip \vs L-knee              \\
        R-hip \vs R-knee              \\
        L-knee \vs L-ankle            \\
        R-knee \vs R-ankle            \\
        L-shoulder \vs L-elbow        \\
        R-shoulder \vs R-elbow        \\
        L-elbow \vs L-wrist           \\
        R-elbow \vs R-wrist           \\
        pelvis \vs L-shoulder         \\
        pelvis \vs R-shoulder         \\
        pelvis \vs neck               \\
        L-hand \vs R-hand             \\
        L-foot \vs R-foot             \\
        \bottomrule
    \end{tabular}
    \end{tabular}
    \vspace{0.2cm}
\label{tab:elementary_posecodes}
\end{table*}

\begin{table*}[t]
\caption{\textbf{Conditions for \posecode categorizations.} The right column provides the condition for a \posecode to have the categorization indicated in the middle column. $v$ represents the estimated value (an angle converted in degrees, or a distance in meters), while the number after the $\pm$ denotes the maximum noise value that can be added to $v$. Thresholds and noise levels depend only on the type of \posecode.}
\centering \scriptsize
\begin{tabular}{lcc}
\toprule
\Posecode type & Categorization & Condition \\
\midrule
\multirow{6}{*}{angle}
    & \texttt{completely bent} & $v \pm 5 \leq 45$ \\
     & \texttt{almost completely bent} &     $45 < v \pm 5 \leq 75$  \\
   &  \texttt{bent at right angle} &   $75 < v \pm 5 \leq 105$  \\
    & \texttt{partially bent} &     $105 < v \pm 5 \leq 135$ \\
    & \texttt{slightly bent} & $135 < v \pm 5 \leq 160$ \\
    & \texttt{straight} & $v \pm 5 > 160$ \\
\midrule 
\multirow{4}{*}{distance}
   & \texttt{close} &  $v \pm 0.05 \leq 0.20$ \\
    &\texttt{shoulder width apart} & $0.20 < v \pm 0.05 \le 0.40$ \\
   & \texttt{spread} &  $0.40 < v \pm 0.05 \le 0.80$ \\
   & \texttt{wide} & $v \pm 0.05 > 0.80$ \\
   \midrule
\multirow{3}{*}{relative position along the X axis}
    & \texttt{at the right of} & $v \pm 0.05 \leq -0.15 $\\
    & \texttt{x-ignored} & $-0.15 < v \pm 0.05 \leq 0.15 $\\
    & \texttt{at the left of} & $v \pm 0.05 > -0.15 $\\
\midrule 
\multirow{3}{*}{relative position along the Y axis}
    &\texttt{below}            & $v \pm 0.05 \leq -0.15 $\\
    &\texttt{y-ignored}  & $-0.15 < v \pm 0.05 \leq 0.15 $\\
    &\texttt{above}            & $v \pm 0.05 > -0.15 $\\
\midrule
\multirow{3}{*}{relative position along the Z axis}
   & \texttt{behind}                  & $v \pm 0.05 \leq -0.15 $\\
    &\texttt{z-ignored}         & $-0.15 < v \pm 0.05 \leq 0.15 $\\
    &\texttt{in front of}                   & $v \pm 0.05 > -0.15 $\\
\midrule
\multirow{3}{*}{pitch \& roll}
    & \texttt{vertical}               & $v \pm 5 \leq 10 $\\
    & \texttt{ignored}       & $10 < v \pm 5 \leq 80 $\\
    & \texttt{horizontal}             & $v \pm 5 > 80 $\\
\midrule
\multirow{2}{*}{ground-contact}
    & \texttt{on the ground} & $v \pm 0.05 \leq 0.10$ \\
    & \texttt{ground-ignored} & $v \pm 0.05 > 0.10$ \\
\bottomrule
\vspace{0.1cm}
\end{tabular} 
\label{tab:elementary_posecodes_thresholds}
\end{table*}
\definecolor{cornflowerblue}{rgb}{0.39, 0.58, 0.93}

\newcommand{\mydot}[1]{
    \begin{tikzpicture}
    \filldraw[fill=#1,draw=#1] circle (3pt);
    \end{tikzpicture}
}

\newcommand{\elep}[3]{\textit{#1}~(#2) = \texttt{#3}}
\newcommand{\superpc}[4]{\multirow{#1}{*}{#2} & \multirow{#1}{*}{\texttt{#3}} & \multirow{#1}{*}{\mydot{#4}}}

\begin{table*}
\caption{\textbf{List of super-\posecodes.} For each super-\posecode, we indicate which body part(s) are subject to description (1st column) and their corresponding pose configuration (each super-\posecode is given a unique category, indicated in the 2nd column). We additionally specify in the 3rd column whether the associated \posecode is skippable for description, following the same color code as for elementary \posecode statistics charts (\protect\mydot{cornflowerblue}: skippable; \protect\mydot{orange}: unskippable). Letters `L' and `R' stand for `left' and `right' respectively. Some super-posecodes can be produced by multiple sets of elementary \posecodes: each set is separated by the word `\textit{or}'.}
\resizebox{0.97\linewidth}{!}{
\centering \scriptsize
\begin{tabular}{lccc}
\toprule
Subject & Configuration & Eligibility & Production \\
\midrule
\superpc{2}{torso}{horizontal}{orange} & \elep{pitch \& roll}{pelvis, L-shoulder}{horizontal} \\
& & & \elep{pitch \& roll}{pelvis, R-shoulder}{horizontal} \\
\midrule
\superpc{5}{body}{bent left}{cornflowerblue} & \elep{relativePos Y}{L-ankle, neck}{below}  \\
& & & \elep{relativePos X}{neck, pelvis}{at left} \\
& & & \textit{or} \\
& & & \elep{relativePos Y}{R-ankle, neck}{below}  \\
& & & \elep{relativePos X}{neck, pelvis}{at left} \\
\midrule
\superpc{5}{body}{bent right}{cornflowerblue} & \elep{relativePos Y}{L-ankle, neck}{below} \\
& & & \elep{relativePos X}{neck, pelvis}{at right} \\
& & & \textit{or} \\
& & & \elep{relativePos Y}{R-ankle, neck}{below} \\
& & & \elep{relativePos X}{neck, pelvis}{at right} \\
\midrule
\superpc{5}{body}{bent backward}{orange} & \elep{relativePos Y}{L-ankle, neck}{below} \\
& & & \elep{relativePos Z}{neck, pelvis}{behind} \\
& & & \textit{or} \\
& & & \elep{relativePos Y}{R-ankle, neck}{below} \\
& & & \elep{relativePos Z}{neck, pelvis}{behind} \\
\midrule
\superpc{5}{body}{bent forward}{cornflowerblue} & \elep{relativePos Y}{L-ankle, neck}{below} \\
& & & \elep{relativePos Z}{neck, pelvis}{front} \\
& & & \textit{or} \\
& & & \elep{relativePos Y}{R-ankle, neck}{below} \\
& & & \elep{relativePos Z}{neck, pelvis}{front} \\
\midrule
\superpc{3}{body}{kneel on left}{orange} & \elep{relativePos Y}{L-knee, R-knee}{below} \\
& & & \elep{ground-contact}{L-knee}{on the ground} \\
& & & \elep{ground-contact}{R-foot}{on the ground} \\
\midrule
\superpc{3}{body}{kneel on right}{orange} & \elep{relativePos Y}{L-knee, R-knee}{above} \\
& & & \elep{ground-contact}{R-knee}{on the ground} \\
& & & \elep{ground-contact}{L-foot}{on the ground} \\
\midrule
\superpc{7}{body}{kneeling}{orange} & \elep{relativePos Y}{L-hip, L-knee}{above} \\
& & & \elep{relativePos Y}{R-hip, R-knee}{above} \\
& & & \elep{ground-contact}{L-knee}{on the ground} \\
& & & \elep{ground-contact}{R-knee}{on the ground} \\
& & & \textit{or} \\
& & & \elep{angle}{L-knee}{completely bent} \\
& & & \elep{angle}{R-knee}{completely bent} \\
& & & \elep{ground-contact}{L-knee}{on the ground} \\
& & & \elep{ground-contact}{R-knee}{on the ground} \\
\midrule
\superpc{2}{hands}{shoulder width apart}{orange} & \elep{distance}{L-hand, R-hand}{shoulder width} \\
& & & \elep{pitch \& roll}{L-hand, R-hand}{horizontal} \\
\midrule
\superpc{2}{feet}{shoulder width apart}{cornflowerblue} & \elep{distance}{L-foot, R-foot}{shoulder width} \\
& & & \elep{pitch \& roll}{L-foot, R-foot}{horizontal} \\
\bottomrule
\end{tabular}}
\label{tab:super_posecodes}
\end{table*}

\begin{figure*}[b]
    \centering
    \includegraphics[width=\linewidth]{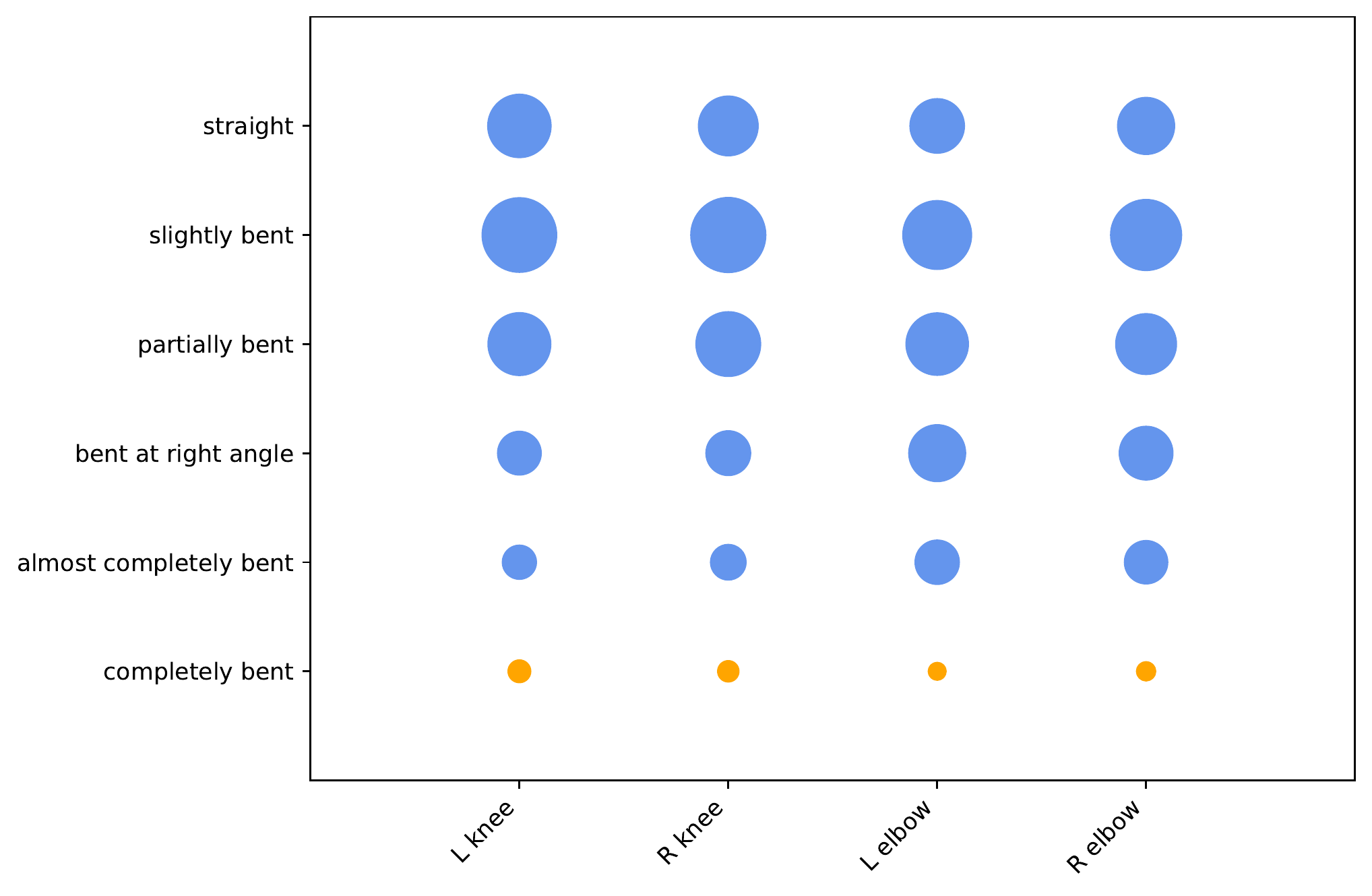}
    \caption{\textbf{Statistics on categorizations of angle \posecodes}, obtained over the poses of \dname-A$_{20}$.
    Letters `L' and `R' refer to left and right body parts respectively. The dot size varies with the proportion of poses that fit to the given categorization. \Posecode categorizations used at captioning time are represented in orange (unskippable) and blue (skippable). For any keypoint, the \posecode interpretation `\texttt{completely bent}' applies to less than 6\% of the poses and is hence defined as unskippable.}
    \label{fig:angle_posecodes}
\end{figure*}

\begin{figure*}[b]
    \centering
    \includegraphics[width=\textwidth]{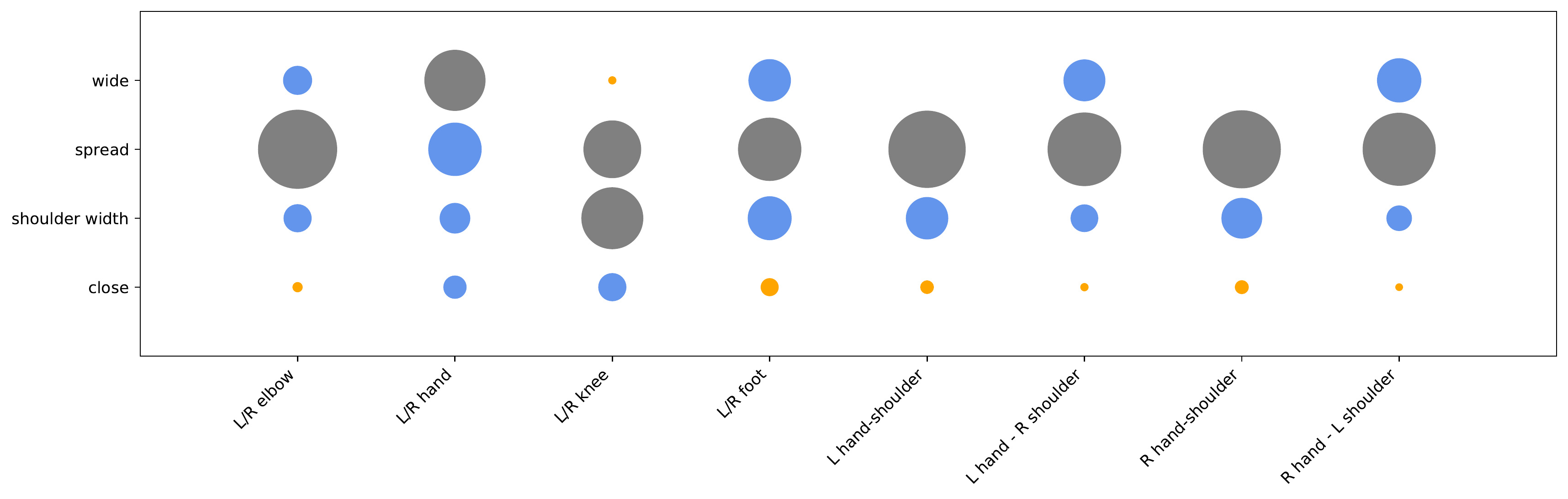}
    \includegraphics[width=\linewidth]{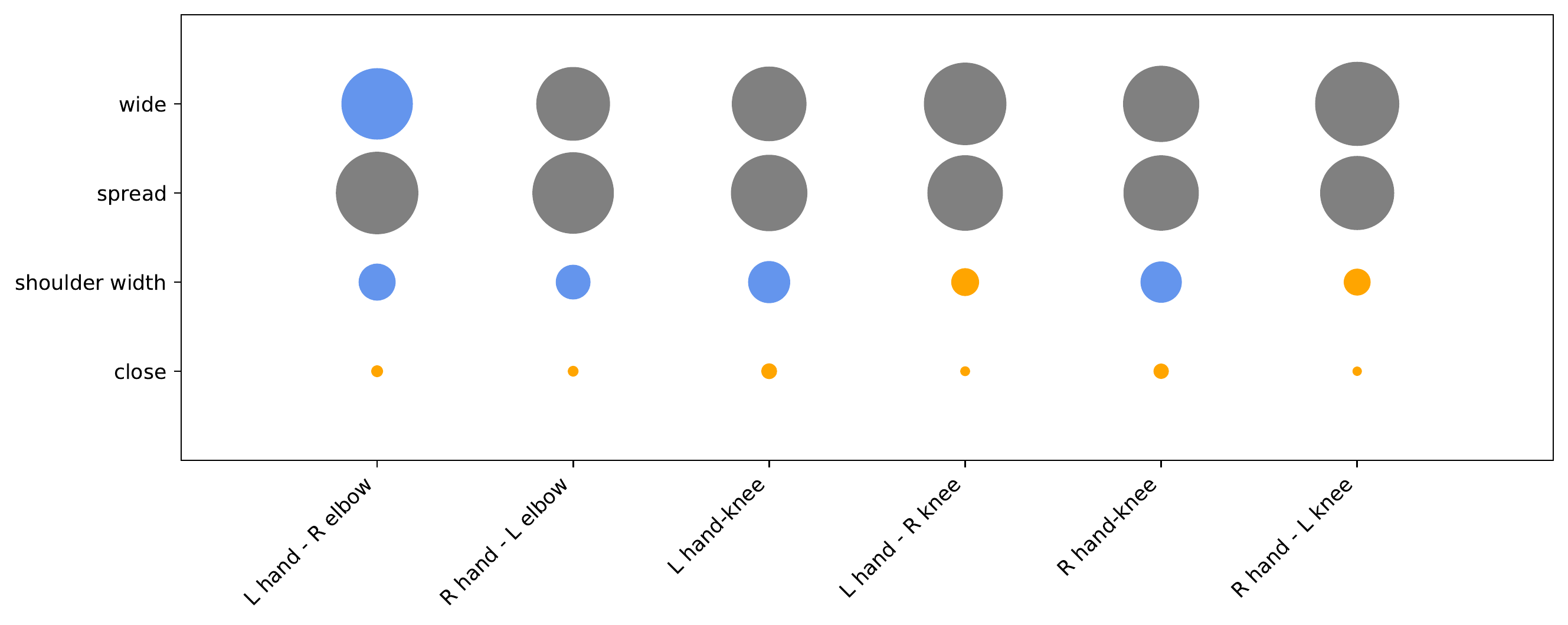}
    \includegraphics[width=\linewidth]{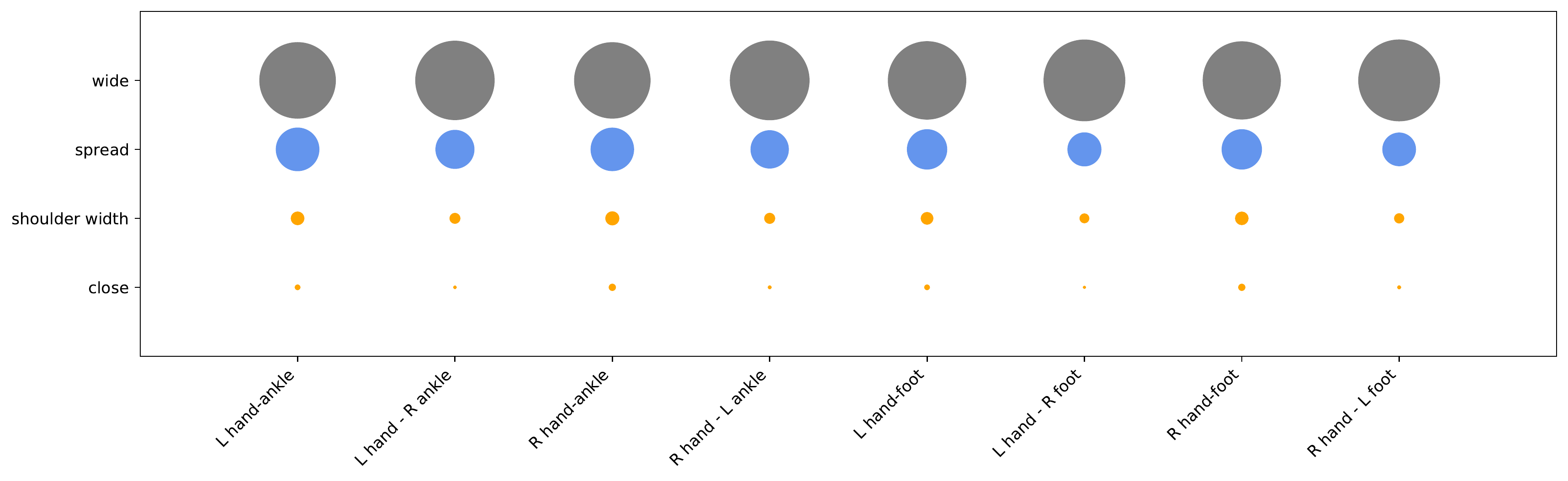}
    \caption{\textbf{Statistics on categorizations of distance \posecodes}, obtained over the poses of \dname-A$_{20}$.
    The first four columns of dots from the top block show distance \posecodes between the left and right corresponding body parts; other columns of dots study the distance between a left or right body part and another left or right body part (when the side of the second body part is not specified, it is the same as for the first body part).
    Letters `L' and `R' refer to left and right body parts respectively. The dot size varies with the proportion of poses that fit to the given categorization.
    The dot color indicates unskippable (orange), skippable (blue), and ignored (grey) \posecodes, based on their scarcity.
    In practice, when a distance \posecode involves one of the hands only, we just consider the `\texttt{close}' categorization.
    }
    \label{fig:distance_posecodes}
\end{figure*}

\begin{figure*}[b]
    \centering
    \includegraphics[width=0.3\linewidth]{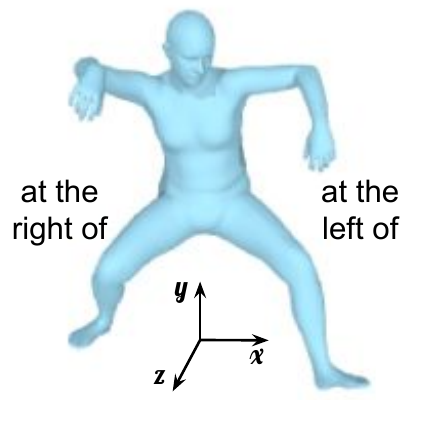}
    \includegraphics[width=\linewidth]{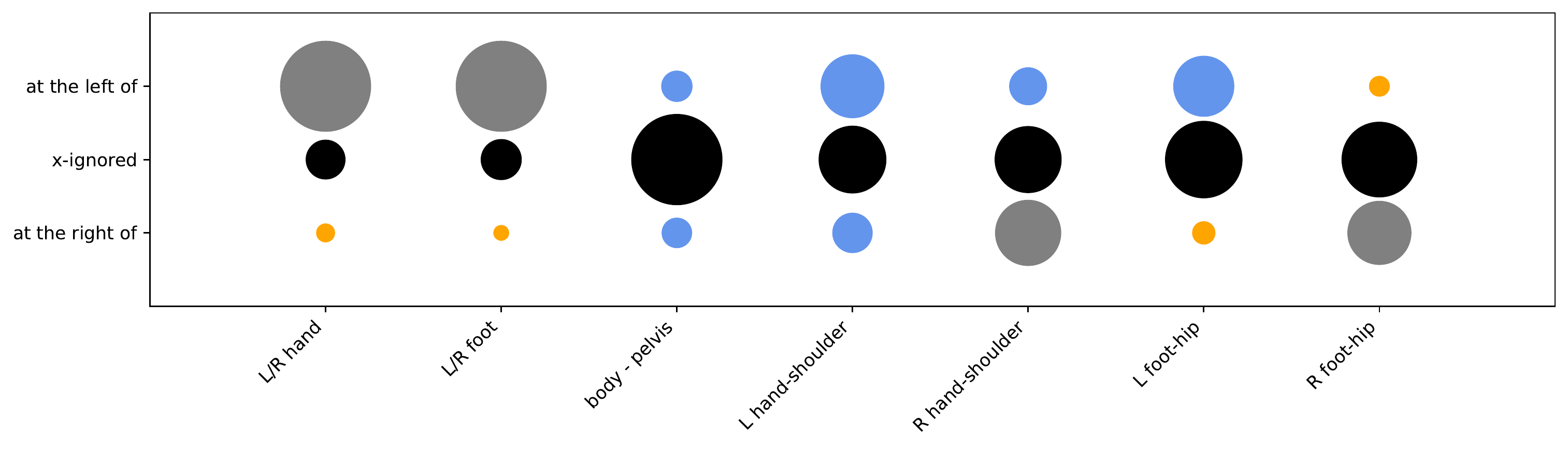}
    \caption{\textbf{Statistics on categorizations of relative position \posecodes along the X axis}, obtained over the poses of \dname-A$_{20}$.
    Letters `L' and `R' refer to left and right body parts respectively. When unspecified, pairs of body parts are from the same side of the body. The dot size varies with the proportion of poses that fit to the given categorization. 
    The dot color indicates unskippable (orange), skippable (blue), and ignored (grey) \posecodes, based on their scarcity. Black dots are ignored because of their inherent ambiguity.
    For instance, it appears that, for less than 6\% of the poses (orange dots), body extremities (hand, foot) are crisscrossed. Such \posecode categorizations are rare, and hence defined as unskippable.
    In some rare cases, dots representing similar relations between left-only body parts and right-only body parts are of different colors (note that dot sizes are still similar) because numbers fall close to the thresholds defining whether a relation should be unskippable/skippable/ignored. In such cases, the same rule is applied for right and left relations, \ie, the left hand (resp. foot) being at the left of the left shoulder (resp. hip) is considered to be a gray dot.
    }
    \label{fig:relposX_posecodes}
\end{figure*}

\begin{figure*}[b]
    \centering
    \includegraphics[width=\linewidth]{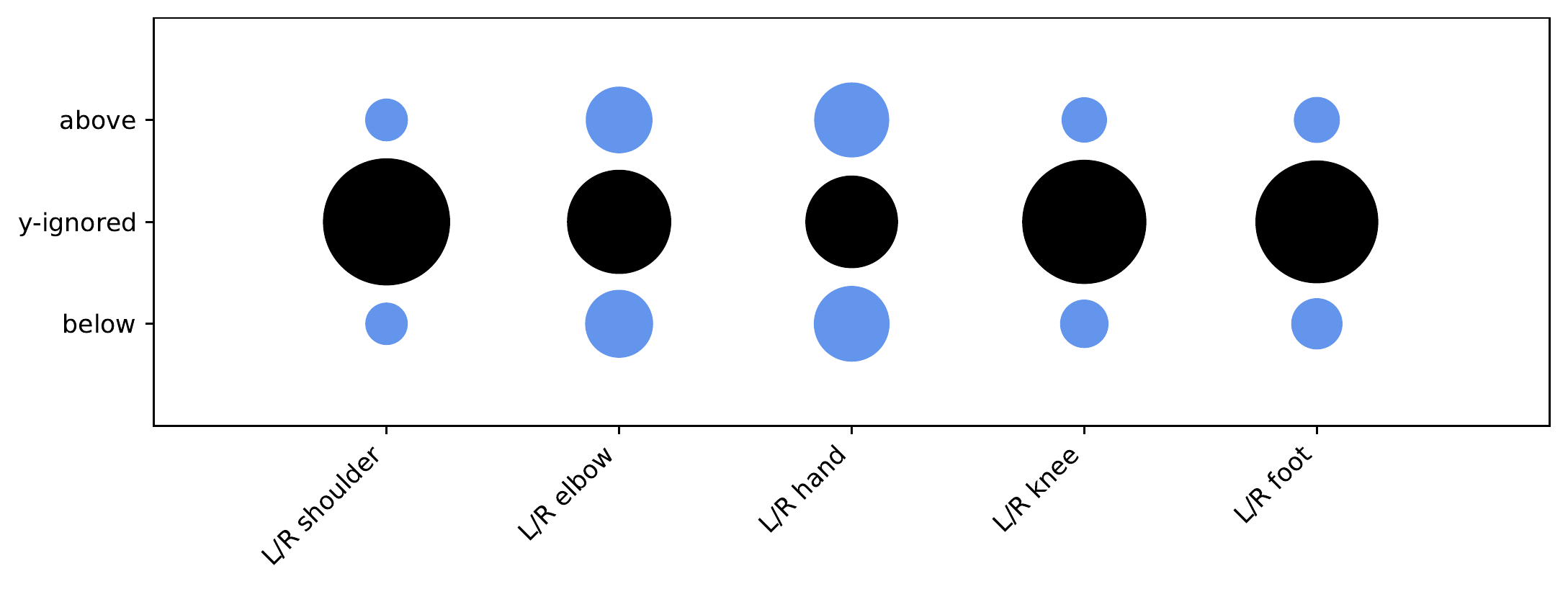}
    \includegraphics[width=\linewidth]{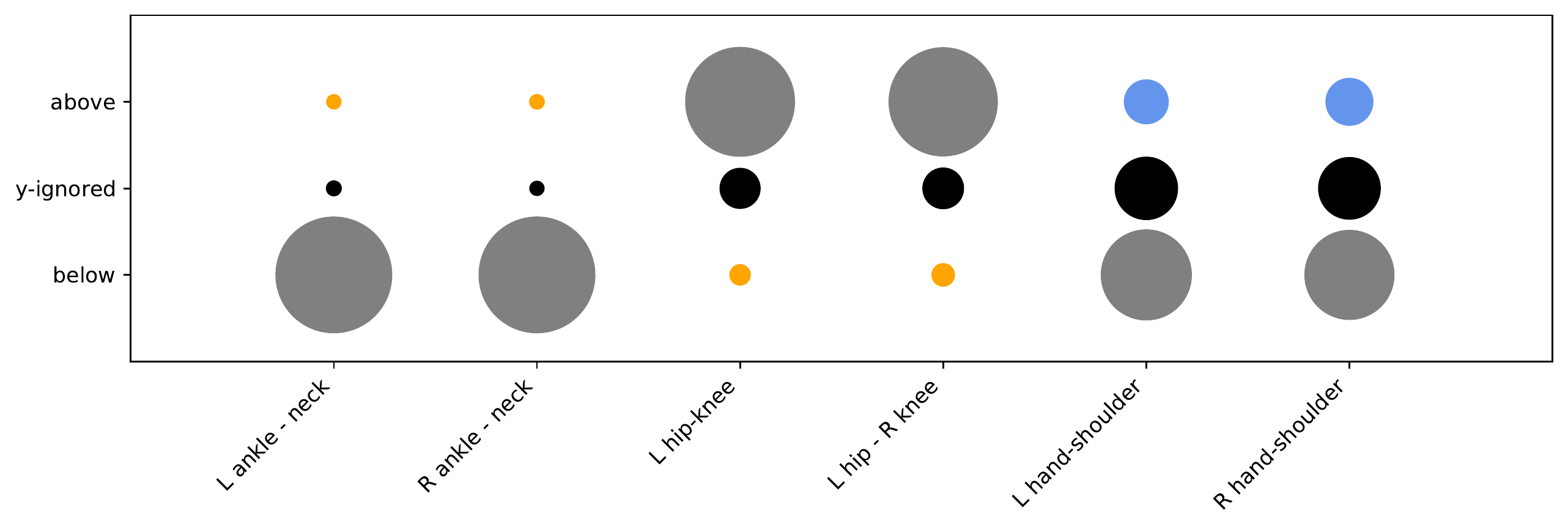}
    \includegraphics[width=\linewidth]{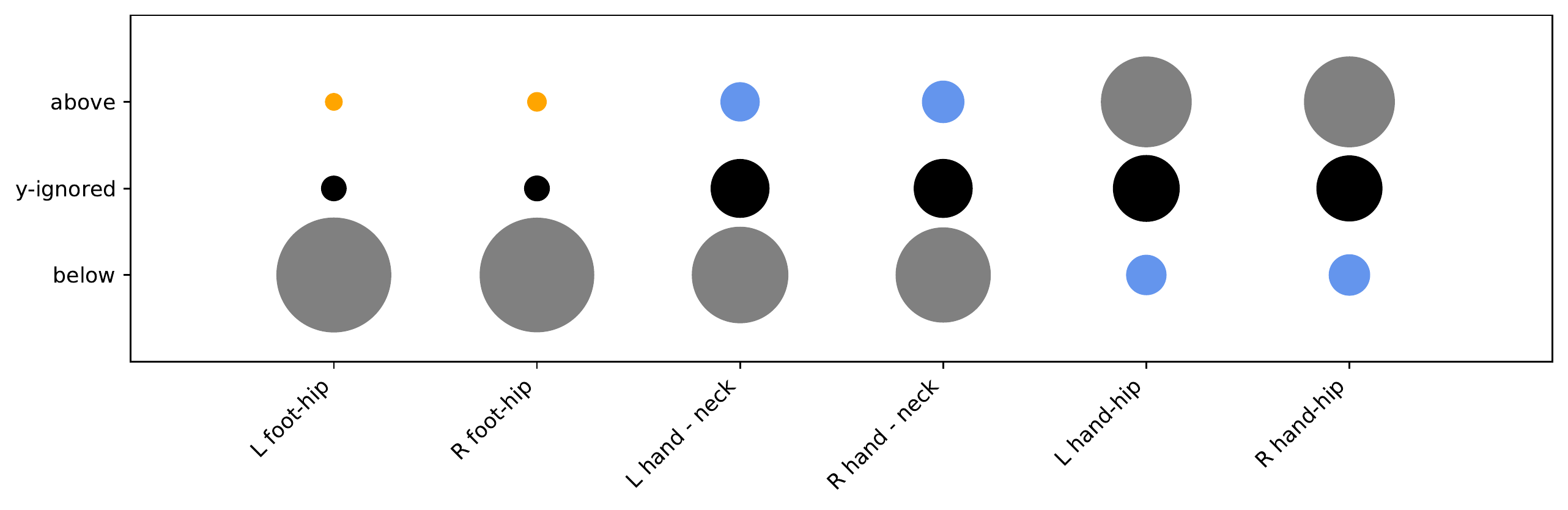}
    \caption{\textbf{Statistics on categorizations of relative position \posecodes along the Y axis}, obtained over the poses of \dname-A$_{20}$.
    The top block shows the relative position of the left body part with respect to the corresponding right body part. Following blocks study other relations; when unspecified, pairs of body parts are from the same side of the body.
    Letters `L' and `R' refer to left and right body parts respectively. The dot size varies with the proportion of poses that fit to the given categorization. 
    The dot color indicates unskippable (orange), skippable (blue), and ignored (grey) posecodes, based on their scarcity. Black dots are ignored because of their inherent ambiguity.
    Note that the dataset is quite balanced regarding left-related and right-related relations (similar dot sizes).
    Some of these posecodes are considered only for super-posecode inference (\eg L ankle - neck); in such cases the scarcity matters less than the provided information.
    }
    \label{fig:relposY_posecodes}
\end{figure*}

\begin{figure*}[b]
    \centering
    \includegraphics[width=\linewidth]{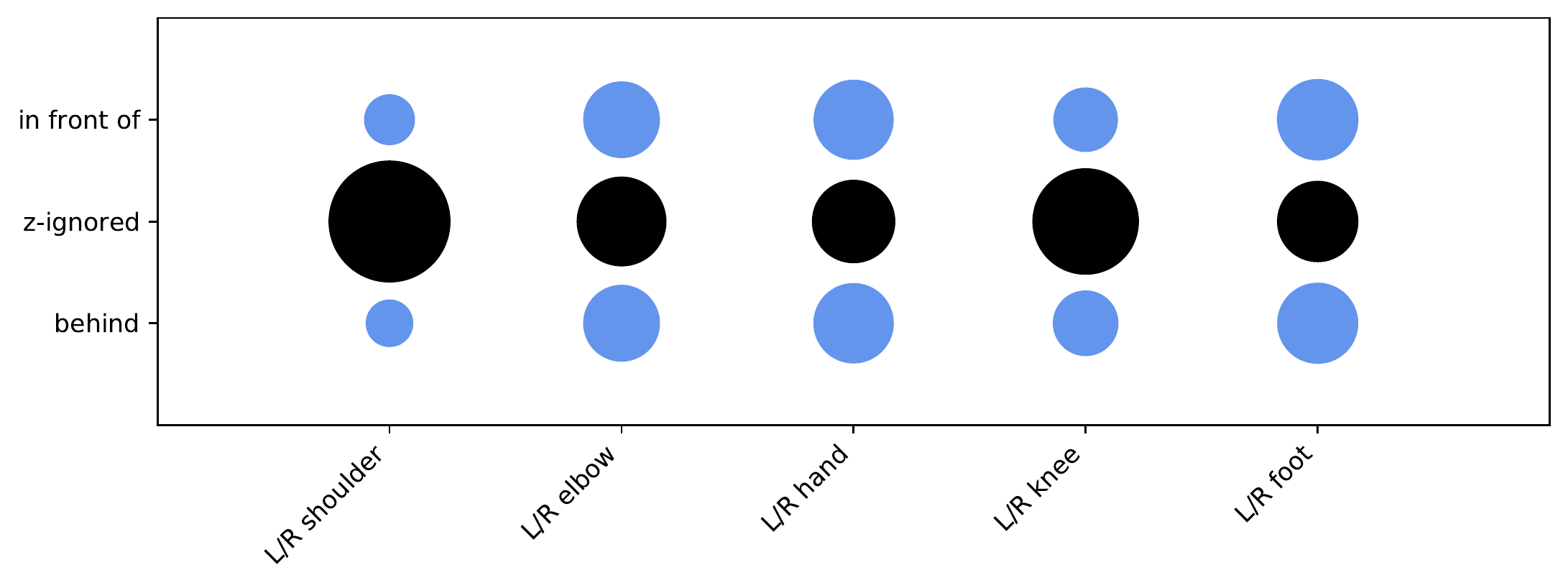}
    \includegraphics[width=\linewidth]{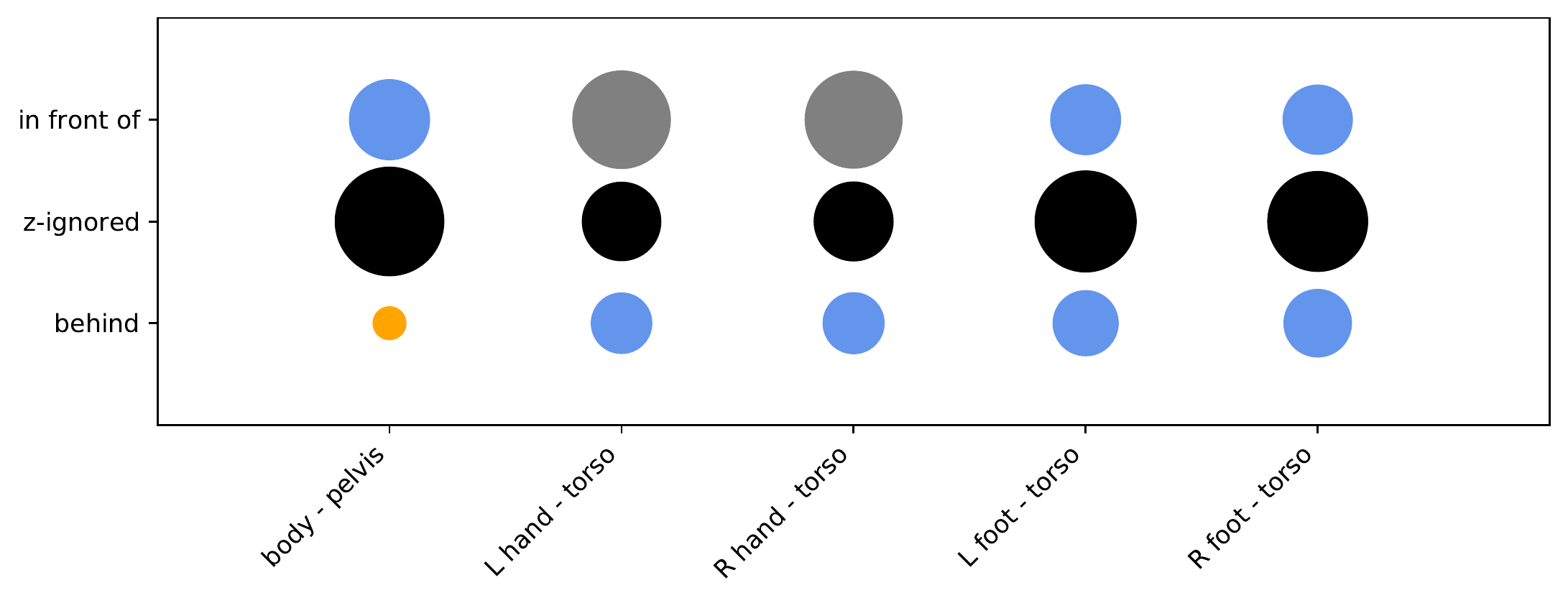}
    \caption{\textbf{Statistics on categorizations of relative position \posecodes along the Z axis}, obtained over the poses of \dname-A$_{20}$.
    The top block shows the relative position of the left body part with respect to the corresponding right body part; the lower block mainly presents the relative position of body extremities (hand/foot) with respect to the torso. The first column of the lower block actually studies the position of the neck with regard to the pelvis to further determine whether the body is bent (forward/backward).
    Letters `L' and `R' refer to left and right body parts respectively. The dot size varies with the proportion of poses that fit to the given categorization. 
    The dot color indicates unskippable (orange), skippable (blue), and ignored (grey) posecodes, based on their scarcity. Black dots are ignored because of their inherent ambiguity.
    }
    \label{fig:relposZ_posecodes}
\end{figure*}

\begin{figure*}[b]
    \centering
    \includegraphics[width=\linewidth]{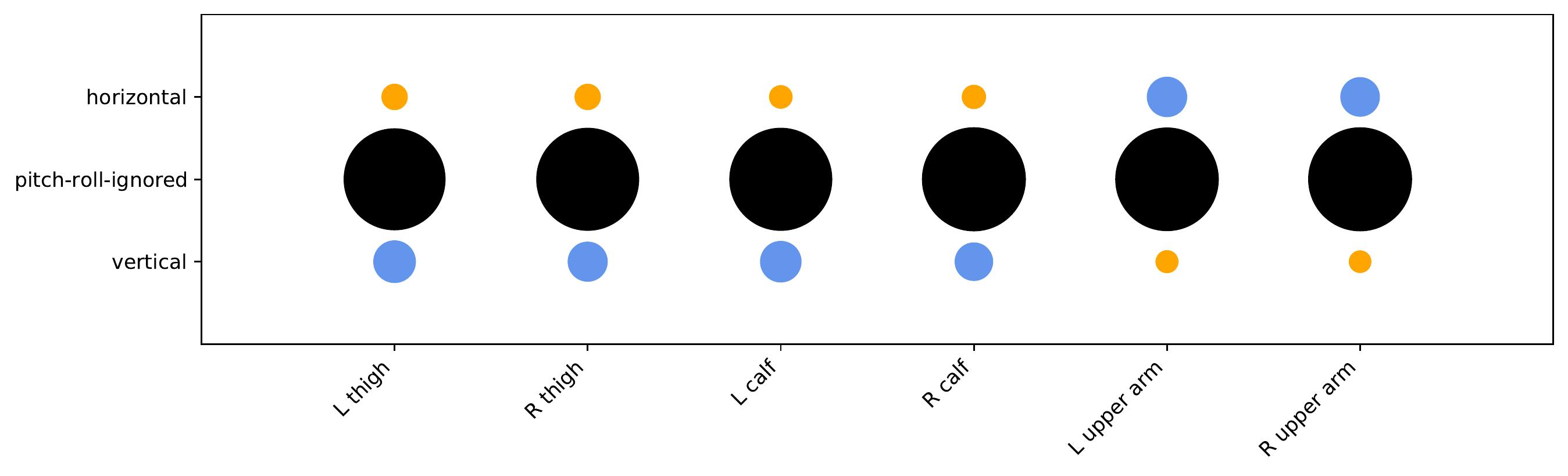}
    \includegraphics[width=\linewidth]{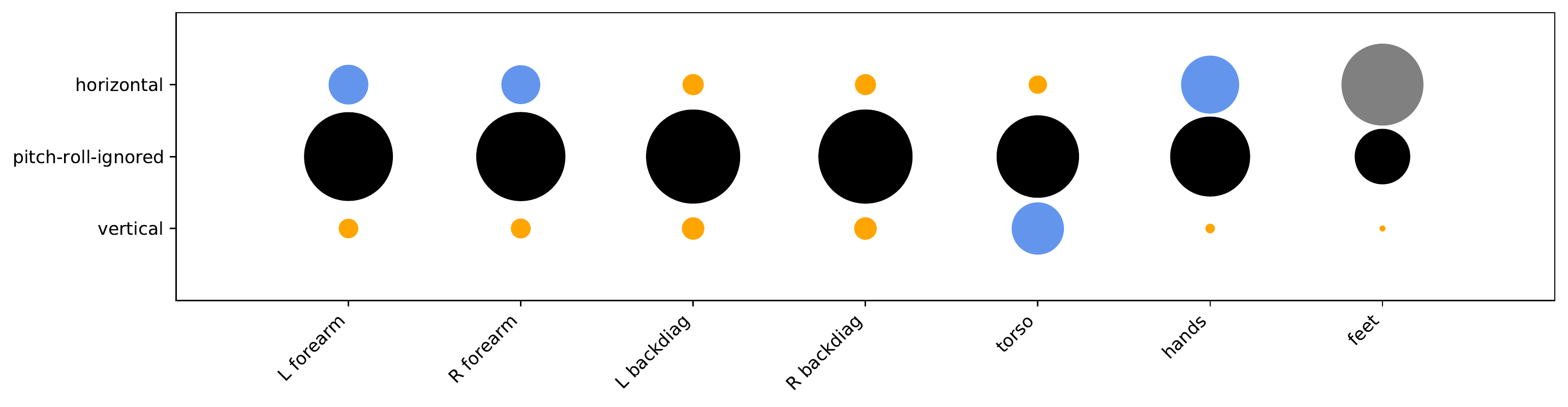}
    \caption{\textbf{Statistics on categorizations of pitch \& roll \posecodes}, obtained over the poses of \dname-A$_{20}$. Letters `L' and `R' refer to left and right body parts respectively. The word `backdiag' refers to the segment between the pelvis and the shoulder, `hands' (resp. `feet') to the segment between the two hands (resp. feet), and `torso' to the segment between the neck and the pelvis.
    The dot size varies with the proportion of poses that fit to the given categorization. 
    The dot color indicates unskippable (orange), skippable (blue), and ignored (grey) posecodes, based on their scarcity. Black dots are ignored because of their inherent ambiguity.
    Some of these posecodes are considered only for super-posecode inference (\eg hands horizontality); in such cases the scarcity matters less than the information provided.
    }
    \label{fig:pitchroll_posecodes}
\end{figure*}

\begin{figure*}[b]
    \centering
    \includegraphics[width=0.8\linewidth]{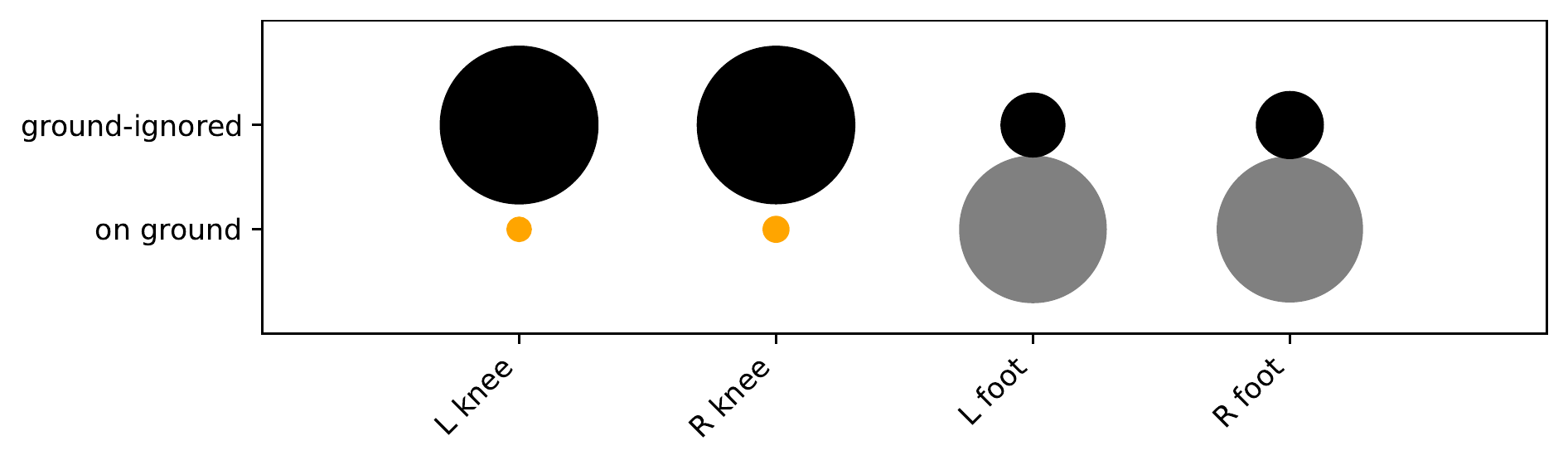}
    \caption{\textbf{Statistics on categorizations of ground-contact \posecodes}, obtained over the poses of \dname-A$_{20}$.
    Letters `L' and `R' refer to left and right body parts respectively. The dot size varies with the proportion of poses that fit to the given categorization. 
    While the dot colors indicate different levels of scarcity, the `\texttt{on the ground}' categorization is used for all of these posecodes independently, for super-posecode inference only.
    }
    \label{fig:groundcontact_posecodes}
\end{figure*}

\myparagraph{\Posecodes statistics.}
In Figures~\ref{fig:angle_posecodes}, \ref{fig:distance_posecodes}, \ref{fig:relposX_posecodes}, \ref{fig:relposY_posecodes}, \ref{fig:relposZ_posecodes}, \ref{fig:pitchroll_posecodes} and \ref{fig:groundcontact_posecodes} we show \posecode statistics obtained over \dname-A$_{20}$. Specifically, circle areas represent the proportion of poses satisfying the corresponding \posecode categorization for the associated keypoints. We use the black and grey colors to denote categorizations that are ignored in the captioning process.
A black circle area means that the corresponding pose configuration is too ambiguous (\eg when the relative distance between two body parts is close to 0, making the detection of the body parts' relative position less obvious.).
Grey circle areas denote trivial pose configurations (\eg when a left body part is at the left of the associated right body part: this is the case most of the time). They correspond to \posecode categorizations that apply to at least 60\% of the poses.
In contrast, \posecode categorizations that describe less than 6\% of the poses are defined as unskippable (\ie, such pose information cannot be randomly discarded during the \posecode selection process), and are colored in orange. All other available \posecodes categorizations, in blue, are skippable (\ie, such pose information can be randomly discarded during the \posecode selection process).
Equivalent information for super-\posecodes is provided in Table~\ref{tab:super_posecodes}.

Most of the time, we follow statistics to consider \posecode categorizations for pose description. In some specific cases, however, we are only interested in a subset of categorizations, and \posecodes were only defined to retrieve such particular body pose information. This was done to infer super-\posecodes later on (as for all ground-contact \posecodes), or to bring in interesting semantics. For instance, distance \posecodes involving one hand and another body part are only considered to inform about the position of the hand via the `\texttt{close}' category; indeed, while someone could describe the right hand as close to the left elbow, they are quite unlikely to point out that the right hand is wide apart from the left elbow. For the sake of completeness, we also present their statistics in the above-mentioned figures.
\section{More about the automatic captioning pipeline}
\label{app:auto_captioning_details}

We provide additional information about some steps of the captioning process.

\myparagraph{Input to the pipeline.}
The process takes 3D joint coordinates of human-centric poses as input. These are inferred using the neutral body shape with default shape coefficients and a normalized global orientation along the y-axis. We use the resulting pose vector of dimension $N\times 3 $ ($N=52$ joints for the SMPL-H model~\cite{mano}), augmented with a few additional keypoints, such as the left/right hands and the torso. They are deduced by simple linear combination of the positions of other joints, and are included to ease retrieval of pose semantics (\eg a hand is in the back if it is behind the torso). Specifically:
\begin{itemize}
    \item the hand keypoint is computed as the center between the wrist keypoint and the keypoint corresponding to the second phalanx of the hand's middle finger.
    \item the torso keypoint is computed as the average of the pelvis, the neck and the third spine keypoint.
\end{itemize}

\myparagraph{What happens to \posecodes contributing to super-\posecodes?}\footnote{\label{footnote1}To reduce the verbosity of this paragraph, we refer to specific \posecode categorizations as `\posecodes'.}
There are three different outcomes for a \posecode that contributes to a super-\posecode:
\begin{itemize}
    \item Some of the elementary \posecodes are only \textit{`support' \posecodes}, and will never make it to the description alone: they only exist for computational purposes and need to be combined with other elementary \posecodes to produce super-\posecodes. For instance, to detect that the torso is parallel to the ground, we check that the two lines between the pelvis and each of the shoulders are horizontal. These two conditions are encoded via `support' \posecodes, which means that if the super-\posecode is not produced because one of the two conditions is not satisfied, the second condition will not be transcribed in the caption: alone, it is meaningless.
    \item Some other \posecodes can be considered as \textit{`semi-support' \posecodes}: they are discarded if the super-\posecode they contribute to is successfully produced, but can make it to the description alone otherwise. For example, one way to detect that the body is kneeling is to check that both knees are completely bent, and in contact with the ground (otherwise the body could be in a squatting position). If all these conditions are met, the body is described as in a kneeling position and there is no need to further precise that the two knees are completely bent. If some of these conditions are not satisfied (\eg the person is standing straight on their right foot), the super-\posecode is not produced, and conversely to a `support' \posecode, the `semi-support' \posecode `the left knee is completely bent' is not discarded, as it carries important information.
    \item Remaining elementary \posecodes, which contribute to super-\posecodes but are neither `support' nor `semi-support' \posecodes will make it to the description, no matter whether the super-\posecodes they contribute to can be produced or not -- unless they are skipped down the road, of course.
\end{itemize}
For more information about which \posecodes are support and semi-support \posecodes, please refer directly to the code.

\myparagraph{How is the redundancy tackled in the captions?}\footnote{To reduce the verbosity of this paragraph, we refer to specific \posecode categorizations as `\posecodes'.}
\Posecodes are numerous, and yet encode a single body pose. Between these constraints and those intrinsic to the human body (\eg arms attached to the torso by the shoulders), information overlap arises quickly.
In the automatic captions, redundancy is tackled in several ways: (1) \posecodes summarized in aggregation rules are removed: information is passed on, not duplicated; (2) most of the \posecodes contributing to super-\posecodes are `support' \posecodes, that exist only for super-\posecode inference and are removed afterwards; (3) redundant \posecodes are further removed thanks to two kinds of ripple effect rules: (i) rules based on statistically frequent pairs and triplets of \posecodes, and (ii) rules based on transitive relations between body parts. In details:
\begin{itemize}
    \item \textbf{Relation-based rules} are mined automatically for each pose, and applied before any aggregation rule. For a given pose, if we have 3 \posecodes telling that $a < b$, $b < c$ and $a < c$ (with $a$, $b$, and $c$ being arbitrary body parts, and $<$ representing a relation of order such as \textit{`below'}), then we keep only the \posecodes telling that $a < b$ and $b < c$, as it is enough to infer the global relation $a < b < c$. For instance, with both \textit{`L hand in front of torso'} and \textit{`R hand behind torso'}, the \posecode \textit{`L hand in front of R hand'} is removed.
    \item \textbf{Statistics-based rules.}
    Let $X$ and $Y$ be two sets of \posecodes. Let's write $p \sim Z$ a pose $p$ that has all \posecodes in a given set $Z$. We define a statistics-based rule $X \Rightarrow Y$ ($X$ `implies' $Y$) if
    \begin{equation}
    \frac{\sum_{p \in \text{PoseScript}} p \sim (X \cup Y)}{\sum_{p \in \text{PoseScript}} p \sim X} \geq \tau ,
    \label{eq:stat_base_def}
    \end{equation}
    with $\tau = 1$ (ideally). In other words, if all the poses which have \posecodes $X \cup Y$ can be summarized as having $X$ only, then any pose that has $X$ necessarily would have $Y$. This is a relatively safe assumption, as poses from PoseScript were selected to be as diverse as possible.
    We automatically mined statistics-based rules $X \Rightarrow Y$ such that $size(X)\leq 2$ and $size(Y)=1$ with the following considerations:
    \begin{itemize}
        \item the rule must involve eligible \posecodes only, \ie, \posecodes that could make it to the description; trivial or ambiguous \posecodes cannot be part of $X$ or $Y$,
        \item the rule must be symmetrically eligible for the left and right sides: the rule must work the same for the whole body,
        \item the rule must affect at least 50 poses, \ie, $\sum_{p \in \text{PoseScript}} p \sim X \geq 50$,
        \item the rule must hold for at least 80\% of the PoseScript poses when $size(X)=2$ (\ie, $\tau = 0.8$) and 70\% when $size(X)=1$ ($\tau = 0.7$).
    \end{itemize}
    Rules were mined based on PoseScript-A$_{20}$.
    We further reviewed all mined rules manually, to keep only the most meaningful and dispose of the following:
    \begin{itemize}
        \item rules where one of the \posecodes in $X$ could be considered an `auxiliary' \posecode, \ie, a \posecode used only to select a smaller set and make the denominator in equation \eqref{eq:stat_base_def} small enough to get past the selection threshold $\tau$. This is particularly obvious when $Y$ and one of the $X$ \posecodes are about the upper body while the other $X$ \posecode is about the lower body, for instance.
        \item rules with weak conditions, \eg when $X$ \posecodes are providing conditions on left body parts relatively to right parts, to derive in $Y$ a `global' result on left body parts.
    \end{itemize}
    Statistics-based rules are computed before but applied after entity-based and symmetry-based aggregation rules; they consist in removing the $Y$ \posecodes if they still exist. For instance, with \textit{`L hand above shoulder'}, \textit{`R hand below hip'}, the \posecode \textit{`L hand above R hand'} is removed. 
\end{itemize}
As a side note, annotators were found to repeat themselves in some captions.

\myparagraph{Entity-based aggregation.}
We defined two very simple entities: the arm (formed by the elbow, and either the hand or the wrist; or by the upper-arm and the forearm) and the leg (formed by the knee, and either the foot or the ankle; or by the thigh and the calf).

\myparagraph{Omitting support keypoints.}
We omit the second keypoint in the phrasing in those specific cases:
\begin{itemize}
    \item a body part is compared to the torso,
    \item the hand is found `above' the head,
    \item the hand (resp. foot) is compared to its associated shoulder (resp. hip), and is found either `at the left of' or `at the right of' of it. For instance, better than having `the right hand is at the left of the left shoulder', which is quite tiresome, we would have \eg `the right hand is turned to the left'.
\end{itemize}

\myparagraph{Limitations of the automatic captioning pipeline.} We discuss here three kinds of limitations.
\begin{itemize}
    \item \textbf{\textit{Rotation-related information.}} As it uses 3D coordinates as input, the automatic captioning pipeline describes relative body part positioning, but does not inform about limb orientation or limb twist. Yet, this could be easily taken into account by designing new posecodes which computation would rely on joint rotations (instead of joint coordinates). This is mostly interesting and worth describing for end limbs like lower arms (\eg to tell whether the inside of the lower arm is facing down or facing up, when the wrist is extended forward). Usually, twists of other limbs either induce visual twists of the end limbs connected to them, or result in different positions of other 3D keypoints. As such, in most of the cases, the captioning of these 3D keypoints does implicitly inform about limb twisting. For instance, twisting the upper arm while keeping the lower arm fixed relatively to it would either increase the twisted appearance of the lower arm (\eg if the arm is straight) or lead to a global change of the 3D position of the wrist (\eg if the elbow is bent). It should be noted that the classification of some other rotations could be integrated as new posecodes, for instance the orientation of the head. We do not claim the designed list of posecodes and super-posecodes to be exhaustive, but the pipeline to be modular and open to new additions. While there will always be room for improvement of the automatic captioning pipeline, the produced automatic captions are only used for pretraining and already prove to be highly beneficial in their current state.
    \item \textbf{\textit{Use of negations.}} We studied the use of negation in human-written captions: less than 5\% of them contain negations. Over 90\% of the times, negation is carried by the word ``not'', as in \eg `[close but] not touching' (22\%), `not quite/fully/completely/very' (19\%) or `not bent' (9\%). Similar negations are easy to integrate in automatic caption templates. We did not include any as the proportion of negations in automatic captions would have been much greater than in human-written captions otherwise. 
    \item \textbf{\textit{Contextual (environment/action) information.}} For \eg pose generation, context can be provided via another modality (\eg an image) or freely expressed in natural language. We include BABEL \cite{babel} action labels in our automatic captions, and annotators were welcome to use analogies in their descriptions, \eg \textit{`as if to climb a ladder'}. We primarily focus on learning explicit fine-grained relations between body parts (detailed \& low-level pose indications). Physical environment constraints are beyond the scope of this work but make for an exciting future research direction.
\end{itemize}
\section{Miscellaneous discussions}
\label{app:misc_discussions}

Some other aspects of this work are further discussed in this section.

\myparagraph{Comparison with CLIP.} We trained our model with pose and text paired data, just like CLIP~\cite{clip} is trained with image and text paired data. Both methods resort to a joint embedding space learned with constrastive learning. The main difference resides in the use of a pose encoder (in our method) instead of an image encoder (in CLIP). 
We test CLIP on our task by rendering the 3D poses under different viewpoints, and doing text-to-image retrieval. We measure a mean recall of 1.6, which is much lower than the 45.3 points obtained by our best model. The CLIP mean recall does not exceed 2, even when testing on the subset of pose descriptions that have less than 77 tokens (size of the CLIP context window) -- this then represents 77\% of the test set, and clears the truncation as the cause for low performance. The measure of the Kozachenko-Leonenko entropy reveals that the text features in CLIP are much closer (-1.32) than for the text features output by model (-0.38), meaning that our model generates more expressive embeddings for pose descriptions. This is not surprising, considering that CLIP is trained on simpler texts and on data (although gigantic) that is not specific to human pose. A qualitative study showed that CLIP works well for very simple instructions such as ``\textit{the person has their arms raised up}" and ``\textit{person sitting}", but fails when asked more detailed and complex descriptions as in PoseScript. This justifies the need for (a) large-scaled text and pose paired data carrying fine-grained pose information, and (b) a model using a 3D pose-specific encoder.

\myparagraph{Pretraining on PoseScript-A.} 
Since the studied tasks require semantic understanding of the pose, using PoseScript-A for pretraining makes a lot of sense: indeed, it pairs poses with texts, which are semantic by nature. 
To emphasize the value of this pretraining, we trained a variational auto-encoder (VAE) model on the poses from PoseScript-A, and used its weights to initialize the pose encoder of our retrieval model from section IV, before finetuning it on PoseScript-H (\texttt{EXP-1}). We also experimented finetuning a retrieval model pretrained on PoseScript-A, where only the weights of the pose encoder would be imported (limiting the initialization to the same layers that could be initialized in the previous experiment); \texttt{EXP-2}. These two models can be compared to a model learned from scratch (no pretraining; \texttt{EXP-0}) and a model whose \textit{both} encoders (the pose's and text's) are initialized with the pretrained weights (\texttt{EXP-3}). Results are presented in Table~\ref{tab:added_results}. It can be observed that any pretraining of the sole pose encoder is useful (+10 points at least on the mRecall, when comparing \texttt{EXP-1} or \texttt{EXP-2} to \texttt{EXP-0}), with the VAE-pretraining \texttt{EXP-1} being better in R@1 and the other ``semantic'' pretraining \texttt{EXP-2} being better in R@50. Eventually, initializing both encoder weights with those of a model pretrained on PoseScript-A is the most beneficial (+78\%, \texttt{EXP-3}), compared to only initializing the pose encoder (+49\%, \texttt{EXP-1}). This highlights the value of PoseScript-A as a set of pose \textit{and} text pairs for pretraining, and thus validates the contribution of the automatic captioning pipeline.

\begin{table}[h]
    \caption{\textbf{Text-to-pose and pose-to-text retrieval results considering different pretraining configurations}, before finetuning on \dname-H. Models all have the GloVe-biGRU configuration. Results are averaged over 3 runs. mRecall is obtained from recalls at 1, 5 and 10.}
    \centering
    \resizebox{\linewidth}{!}{%
    \begin{tabular}{llccccc@{}}
    \toprule
    \multirow{2}{*}{\texttt{EXP}} & \multirow{2}{*}{Pretraining} &  \multirow{2}{*}{mRecall\color{OliveGreen}{$\uparrow$} } & \multicolumn{2}{c}{pose-to-text} & \multicolumn{2}{c}{text-to-pose} \\
    \cmidrule(lr){4-5} \cmidrule(lr){6-7}
    & & & $R@1$\color{OliveGreen}{$\uparrow$} & $R@10$\color{OliveGreen}{$\uparrow$} & $R@1$\color{OliveGreen}{$\uparrow$} & $R@10$\color{OliveGreen}{$\uparrow$} \\
    \midrule
    \texttt{0} & none & 23.0 \tiny{${\pm}$ 0.6} & 8.9 \tiny{${\pm}$ 0.3} & 34.8 \tiny{${\pm}$ 0.5} & 9.3 \tiny{${\pm}$ 1.0} & 35.7 \tiny{${\pm}$ 0.9} \\
    \texttt{1} & pose encoder (via VAE) & 34.2 \tiny{${\pm}$ 0.2} & 15.5 \tiny{${\pm}$ 0.7} & 48.7 \tiny{${\pm}$ 0.5} & 15.5 \tiny{${\pm}$ 0.5} & 50.2 \tiny{${\pm}$ 0.1} \\
    \texttt{2} & pose encoder & 33.8 \tiny{${\pm}$ 1.1} & 14.2 \tiny{${\pm}$ 1.2} & 49.1 \tiny{${\pm}$ 1.7} & 13.8 \tiny{${\pm}$ 0.4} & 50.9 \tiny{${\pm}$ 0.8} \\
    \texttt{3} & pose \& text encoders & 40.9 \tiny{${\pm}$ 0.1} & 19.8 \tiny{${\pm}$ 0.4} & 56.2 \tiny{${\pm}$ 0.7} & 19.9 \tiny{${\pm}$ 0.6} & 57.9 \tiny{${\pm}$ 0.3} \\
    \bottomrule
    \end{tabular}
    }
    \label{tab:added_results}
\end{table}

\end{document}